\pdfoutput=1
\documentclass[sigconf]{acmart}

\usepackage{url}
\usepackage{amsmath}
\usepackage{algorithm}
\usepackage{algorithmicx}
\usepackage{algpseudocode}
\usepackage{mathrsfs}
\usepackage{amsmath}
\usepackage{multirow}
\usepackage{subfigure}

\usepackage{amssymb}
\usepackage{bm}
\usepackage{caption}
\usepackage{enumitem}

\usepackage{makecell}

\copyrightyear{2025}
\acmYear{2025}
\setcopyright{cc}

\setcctype{by}
\acmConference[WWW '25]{Proceedings of the ACM Web Conference 2025}{April
28-May 2, 2025}{Sydney, NSW, Australia}
\acmBooktitle{Proceedings of the ACM Web Conference 2025 (WWW '25), April
28-May 2, 2025, Sydney, NSW, Australia}
\acmDOI{10.1145/3696410.3714514}
\acmISBN{979-8-4007-1274-6/25/04}

\begin{document}

\title{MedGNN: Towards Multi-resolution Spatiotemporal Graph Learning for Medical Time Series Classification}

\author{Wei Fan}
\authornote{Both authors contributed equally to this research.}
\email{weifan.oxford@gmail.com}
\affiliation{%
  \institution{University of Oxford}
  \city{Oxford}
  \country{UK}
}

\author{Jingru Fei}
\authornotemark[1]
\email{jingrufei@bit.edu.cn}
\affiliation{%
  \institution{Beijing Institute of Technology}
  \city{Beijing}
  \country{China}
}

\author{Dingyu Guo}
\email{kevguo@uchicago.edu}
\affiliation{%
  \institution{University of Chicago}
  \city{Chicago}
  \country{US}
}

\author{Kun Yi}
\authornote{Corresponding authors.}
\email{kunyi.cn@gmail.com}
\affiliation{%
  \institution{State Information Center of China}
  \institution{North China Institute of Computing Technology}
  \city{Beijing}
  \country{China}
}

\author{Xiaozhuang Song}
\email{xiaozhuangsong1@link.cuhk.edu.cn}
\affiliation{%
  \institution{The Chinese University of Hong Kong, Shenzhen}
  \city{Shenzhen}
  \country{China}
}

\author{Haolong Xiang}
\email{hlxiang@nuist.edu.cn}
\affiliation{%
  \institution{Nanjing University of Information Science and Technology}
  \city{Nanjing}
  \country{China}
}

\author{Hangting Ye}
\email{yeht2118@mails.jlu.edu.cn}
\affiliation{%
  \institution{Jilin University}
  \city{Changchun}
  \country{China}
}

\author{Min Li}
\authornotemark[2]
\email{limin@mail.csu.edu.cn}
\affiliation{%
  \institution{Central South University}
  \city{Changsha}
  \country{China}
}



\begin{abstract}
  Medical time series has been playing a vital role in real-world healthcare systems as valuable information in monitoring health conditions of patients. Accurate classification for medical time series, e.g., Electrocardiography (ECG) signals, can help for early detection and diagnosis, thus improving patient outcomes and the quality of life. Traditional methods towards medical time series classification rely on handcrafted feature extraction and statistical methods; with the recent advancement of artificial intelligence, the machine learning and deep learning methods have become more popular. However, existing methods often fail to fully model the complex spatial dynamics under different scales, which ignore the dynamic multi-resolution spatial and temporal joint inter-dependencies. Moreover, they are less likely to consider the special baseline wander problem as well as the multi-view characteristics of medical time series, which largely hinders their prediction performance. To address these limitations, we propose a Multi-resolution Spatiotemporal Graph Learning framework, \textit{{MedGNN}}, for medical time series classification. Specifically,  we first propose to construct multi-resolution adaptive graph structures to learn dynamic multi-scale embeddings. Then, to address the baseline wander problem, we propose Difference Attention Networks to operate self-attention mechanisms on the finite difference for temporal modeling. Moreover, to learn the multi-view characteristics, we utilize the Frequency Convolution Networks to capture complementary information of medical time series from the frequency domain. In addition, we introduce the Multi-resolution Graph Transformer architecture to model the dynamic dependencies and fuse the information from different resolutions. Finally, we have conducted extensive experiments on multiple medical real-world datasets that demonstrate the superior performance of our method. Our Code is available at this repository: \url{https://github.com/aikunyi/MedGNN}.

\end{abstract}

\vspace{-2mm}

\vspace{-2mm}

\maketitle

\vspace{-2mm}
\section{Introduction}
Medical time series data, such as Electroencephalography (EEG) and Electrocardiography (ECG) signals, has been playing a vital role in real-world healthcare systems by providing valuable information in monitoring health conditions of patients. 
EEG signals, which measure the electrical activity of the brain, are widely used to diagnose and monitor various neurological disorders, including epilepsy, Alzheimer's disease, and sleep disorders \cite{cohen2017does}. Similarly, ECG signals, which record the electrical activity of the heart, are essential for diagnosing and monitoring cardiovascular diseases, such as arrhythmias, myocardial infarction, and congestive heart failure \cite{berkaya2018survey}. Classifying these medical time series is of paramount importance as it could enable the early detection of abnormalities, accurate diagnosis, and personalized treatment. By identifying patterns and features indicative of specific conditions, medical time series classification can assist clinicians in making timely diagnoses \cite{ismail2019deep} and facilitate adapting treatment plans accordingly, potentially leading to improved patient outcomes and quality of life.

Traditionally, medical time series classification has been primarily relied on handcrafted feature extraction, which often involve domain expertise to identify relevant features from the raw data. For instance, for ECG analysis, many features such as R-peak amplitude and heart rate variability would be mannually extracted~\cite{berkaya2018survey}.  
Later, statistical methods have also been applied to medical time series classification: the autoregressive models, hidden Markov models, and Gaussian mixture models have been used to capture the temporal dependencies and dynamics in ECG and EEG signals~\cite{schaffer2021interrupted,vincent2010spatially,turner2020design}. Though statistical methods can provide robust results and handle uncertainty, they often make strong assumptions about the data distribution and may struggle with complex, non-linear patterns. 
With the advent of artificial intelligence, various deep learning methods have been applied to medical time series classification~\cite{morid2023time}: convolution neural networks have been particularly successful in learning representations directly from raw time series, such as EEGNet~\cite{eegnet_2016}; Transformer-based methods have been applied into the medical time series classification~\cite{medformer_2024}. In addition, graph neural network has also been adopted for multivariate time series classification~\cite{ZhaLZH22,YounisHA24}.

However, these methods often fail to fully model the complex spatial (channel) dynamics under different scales, which ignore the dynamic multi-resolution spatial and temporal joint interdependencies. Moreover, most of them are usually for general classification, without considering the special problem, such as baseline wander, as well as the multi-view characteristics in medical time series, which largely hinders their prediction performance.
To address these limitations, we aim to propose a novel framework to learn the multi-scale and multi-view representations for medical time series classification. 
However, several challenges arise in achieving this goal: 
(i) \textit{{how to model the dynamic spatial structures between different time series channels with multiple resolutions?}} Since the medical time series usually includes multiple channels, the dynamic spatial dependencies keep changing with the resolutions or scales of time series, which need to be properly modeled for accurate classification;
(ii) \textit{how to learn the the multi-view characteristics of medical time series while addressing the baseline wander problem?} The baseline wander problem~\cite{blanco2008ecg}, i.e., the constant offsets or slow drifts towards baseline measurements of medical series would always hinders the models in learning the key patterns and fluctuations; meanwhile the multi-view characteristics of medical time series based on the features from both the time domain and the frequency domain are usually ignored, largely hindering the classification.

To tackle these challenges, we introduce a Multi-resolution Spatiotemporal Graph Learning framework, \textit{{MedGNN}}.
Specifically, in our MedGNN framework, we first propose to construct multi-resolution adaptive graph structures to learn dynamic spatial temporal representations, where we utilize different kernels of convolutions to extract multi-scale medical time series embeddings to cover the local and global patterns. We construct multi-resolution graphs based on the learned embeddings to model the dynamic spatial dependencies among different channels; the graph structures are adaptively learned to reflect the changing correlations at different resolutions. 
Then, to address the baseline wander problem and learn the multi-view characteristics, we propose two novel networks for temporal modeling: (i) the Difference Attention Networks focus on the temporal changes in the medical time series, which operates self-attention machenisms on the finite difference (e.g., first-order difference) along the temporal dimension, targeting capturing key temporal patterns while mitigating the impact of baseline wander; (ii) the Frequency Convolution Networks captures complementary information in the frequency domain by applying Fourier transform and frequency-domain convolutions~\cite{yi2024frequency, yi2024filternet}, providing a multi-view perspective of the temporal dynamics for medical time series.
In addition, to learn the complicated multi-resolution spatiotemporal graph representations, we utilize the Multi-resolution Graph Transformer architecture to model the dynamic spatial dependencies and fuse the information from different resolutions. Our main contributions are mainly as follows:
\vspace{-0mm}
\begin{itemize}[leftmargin=*]
    \item We propose a novel approach for medical time series classification to capture the complex multi-view spatiotemporal dependencies and multi-scale dynamics of medical time series through multi-resolution learning.
    \item We construct adaptive graph structures at different resolutions to model spatial correlations among time series channels and utilize a Multi-resolution Graph Transformer architecture for the resolution learning and information fusion.
    \item We propose Difference Attention Network and Frequency Convolution Network, for temporal modeling to overcome baseline wander problem of medical time series and meanwhile capture multi-view characteristics from the time and frequency domain.
    \item We have conducted extensive experiments on multiple medical time series datasets, including both ECG and EEG signals, to demonstrate the superior performance of our proposed framework compared to state-of-the-art methods, highlighting its great potential for the real-world clinical applications.
\end{itemize}

\vspace{-2mm}
\section{Related Work}

\subsection{Medical Time Series Classification}
Time series classification is a crucial yet challenging problem in the field of data mining, as it involves identifying patterns in sequential data over time~\cite{tsc_survey_2019}. 
Medical time series, as a specialized form of time series data collected from human physiological signals, such as EEG~\cite{YangW023} and ECG~\cite{KiyassehZC21}, present unique challenges and opportunities~\cite{JafariSKBSLGA23}. Continuously analyzing medical time series, especially as new conditions or classes of data emerge, is essential for health monitoring~\cite{WangSY23} and making informed medical decisions~\cite{AdmassB24}, highlighting the importance of medical time series classification.

Traditionally, one of the most widely used approaches for medical time series classification has been the nearest neighbor (NN)~\cite{LinesB15} classifier, often combined with distance measures such as dynamic time warping (DTW)~\cite{BagnallLBLK17} or shapelet-based methods~\cite{BostromB17}. These techniques have demonstrated effectiveness in various applications due to their simplicity and interpretability. Later statistical models such as the autoregressive models~\cite{schaffer2021interrupted} and Gaussian mixture models~\cite{vincent2010spatially} have been used to capture the medical time series.
In recent years, deep learning methods have significantly advanced the field of medical time series classification. For example, EEGNet~\cite{eegnet_2016} introduced the use of depthwise and separable convolutions to develop a model specifically designed for EEG data, capturing essential EEG feature extraction techniques for brain-computer interfaces (BCI). COMET~\cite{comet_2023} proposed a hierarchical contrastive representation learning framework that is tailored to the unique characteristics of medical time series data. Medformer~\cite{medformer_2024} introduced a multi-granularity patching transformer architecture that addresses the specific challenges of medical time series classification, providing a specialized solution for capturing complex temporal patterns.

\vspace{-2.5mm}
\subsection{Graph Neural Networks for Time Series}
Graph neural networks (GNNs) have shown promising performance in time series analysis due to their ability to capture complex dependencies between time-series variables~\cite{gnn_survey_2023}. By representing data as a graph, GNNs can effectively model relationships across different variables and time steps. 
Some GNN-based models, such as STGCN~\cite{YuYZ18} and DCRNN~\cite{LiYS018}, rely on a pre-defined graph structure, which is often unavailable or difficult to determine in many real-world scenarios. To address this limitation, recent research has focused on learning graph structures directly from the data, enabling automatic modeling of the topological relationships among variables. AGCRN~\cite{Bai2020nips} enhances graph convolutional networks through a data-adaptive graph generation module and a node-adaptive parameter learning module. MTGNN~\cite{wu2020connecting} introduces an effective approach to learn and exploit the inherent dependencies among variables. FourierGNN~\cite{yi2024fouriergnn} captures frequency domain spatial-temporal correlations.
More recent models have continued to push the boundaries of GNN-based time series analysis. RainDrop~\cite{ZhangZTZ22} introduces a GNN framework designed to handle irregularly sampled and multivariate time series, learning the dynamics of sensors directly from observational data without requiring any prior knowledge of the relationships. SimTSC~\cite{ZhaLZH22} presents a simple yet general framework that uses GNNs to model similarity information, which helps improve time series classification by leveraging similarity patterns across different time points and variables. MTS2Graph~\cite{YounisHA24} offers a strategy by constructing a graph that captures the temporal relationships between extracted patterns at each layer of the network. 
These methods underscore the development in GNN-based methods, highlighting their potential to revolutionize time series classification by effectively capturing complex dependencies and adapting to diverse scenarios.



\vspace{-2mm}
\section{Problem Formulation}



Given a collection of medical time series from $N$ participants denoted by $\{P_1, \cdots, P_N \}$, each participant has multiple samples of collected series $\{\mathbf{X}^{1}_{P_n}, \cdots, \mathbf{X}^{k_n}_{P_n}\}$, where $k_n$ is the number of samples for the $n$-th participant. Each data sample $\mathbf{X}^{i}_{P_n} \in \mathbb{R}^{T \times C}$ represents the collected multivariate medical time series (e.g., multi-lead ECG) in one participation, where $T$ denotes the number of timestamps and $C$ is the number of channels; the corresponding label for sample $\mathbf{X}^{i}_{P_n}$ is represented as $y^{i}_{P_n} \in \{0, 1\}$ for binary medical classification problems where  0 indicates a healthy participant and 1 indicates a participant diagnosed with a specific disease, or represented as $y^{i}_{P_n} \in \{1,2, \cdots, c\}$ for multi-class medical classification problems where each class corresponds to one kind of diseases or conditions.
The objective of medical time series classification problem is to learn a mapping function 
$f: \mathbb{R}^{T*C} \rightarrow \mathbb{R}^1$
that can accurately predict the label based on medical time series samples in each participation. Formally, given a participant's time series sample $\mathbf{X}^{i}_{P_n}$, the goal is to predict the corresponding label $y^i_{P_n}$ to indicate the disease or condition, which can be written as:
\begin{equation}
\hat{y}^i_{P_n} = f(\mathbf{X}^{i}_{P_n}; \theta),
\end{equation}
where $\hat{y}^i_{P_n}$ is label for $i$-th sample for $n$-th participant; the mapping function $f$ is parameterized by the learnable parameters $\theta$.

To ensure the real-world clinical utility of the model, it is important to evaluate the generalization ability on unseen participants. Thus we split the dataset based on participants: specifically, we let $\mathcal{P}{{train}}$, $\mathcal{P}{{val}}$, and $\mathcal{P}{{test}}$ denote the disjoint sets of participants used for training, validation, and testing, respectively. With training and testing are on different participants, the generalization of the developed model can be evaluated on unseen participants or patients, which can simulate a more realistic estimate of its potential performance in real-world clinical settings. 
In addition, since one patient may visit hospitals for tests for many times, we also consider a complementary settings that split the dataset into training, validation and test data only relying on individual samples that can be represented by $\mathcal{X}_{train}, \mathcal{X}_{val}$ and $\mathcal{X}_{test}$ as disjoint sets of samples.


\vspace{-1mm}
\section{Methodology}
\begin{figure*}[!t]
    \centering
    \includegraphics[width=0.94\linewidth]{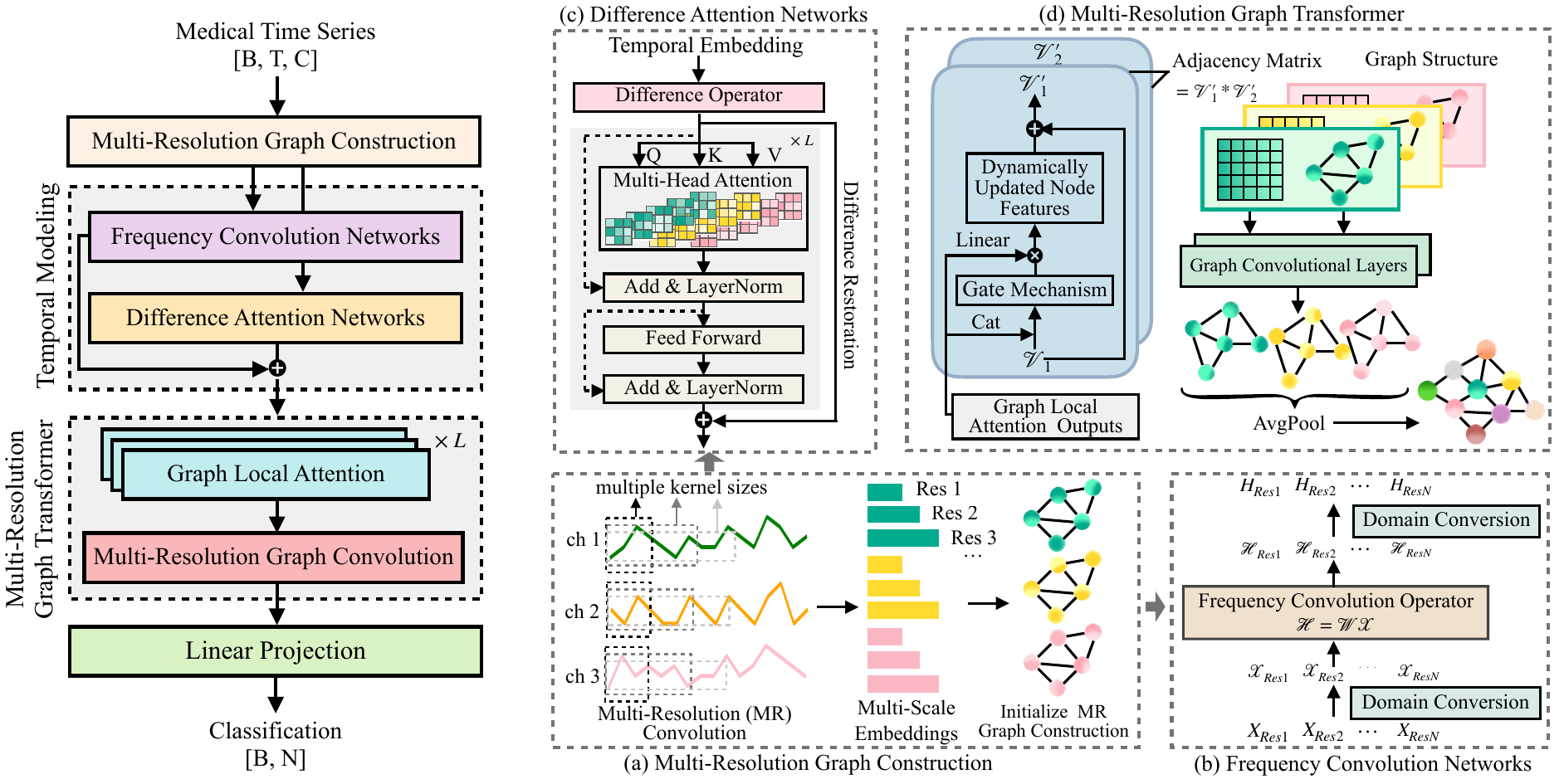}
    \vspace{-2mm}
    \caption{The overall architecture of MedGNN. (a) Multi-resolution graph construction is utilized to learn the dynamic spatiotemporal representations. (b) Frequency convolution networks are applied to provide a multi-view perspective of the temporal dynamics by applying convolutions in the frequency domain. (c) Difference attention networks are employed to capture key temporal patterns while mitigating the impact of baseline wander. (d) Multi-resolution graph transformer is leveraged to model the dynamic spatial dependencies and and fuse the information from different resolutions.}
    \label{fig:framework}
    \vspace{-2mm}
\end{figure*}

To address the aforementioned challenges of medical time series classification, in this section, we elaborate on our proposed \textit{MedGNN}, a Multi-resolution Spatiotemporal Graph Learning framework.
The overall architecture of MedGNN is illustrated in Figure \ref{fig:framework}. It mainly consists of multi-resolution graph construction, difference attention networks, frequency convolution networks and multi-resolution graph transformer. For the medical time series, multi-resolution graph construction is utilized to learn the dynamic spatiotemporal representations, and then difference attention networks and frequency convolution networks are designed to capture the comprehensive temporal dynamics. Finally, multi-resolution graph transformer is employed to model the dynamic spatial dependencies from different resolutions.
In the following, 
we elaborate on the core components of MedGNN from Section \ref{sec:1} to Section \ref{sec:4}.

\vspace{-2mm}
\subsection{Multi-resolution Graph Construction with Multi-scale Embedding Learning}\label{sec:1}

Medical time series typically consist of multiple channels that are often closely correlated. For example, in EEG signals, different brain regions may exhibit synchronization patterns, indicating real functional connectivity in the brain~\cite{teplan2002fundamentals}; in ECG signals, different leads can provide complementary information about the cardiac electrical activity~\cite{berkaya2018survey}. This motivates us to construct explicit structures to model channel-level dynamics and temporal dynamics for medical time series. To this end, we propose a novel Multi-resolution Adaptive Graph Structure Learning approach to model different levels of dynamics. This approach mainly consists of two main steps: 1) learning multi-resolution embeddings, and 2) constructing multi-resolution graphs based on the learned embeddings. 
\subsubsection{Multi-scale Embedding Learning}
Given a medical time series sample $\mathbf{X}^{i}_{P_n} \in \mathbb{R}^{T \times C}$, we aim to learn multi-scale temporal embeddings that capture the local patterns and dynamics at different time scales. To achieve this, we employ a set of 1-\textit{d} convolution networks for different resolutions. Let ${\mathbf{K}_1, \cdots, \mathbf{K}_M}$ denote a series of $M$ kernel sizes, where each kernel $\mathbf{K}_m \in \mathbb{R}^{k_m \times 1}$ has a size of $k_m$ along the time dimension for different channels.
The multi-scale embeddings are obtained by applying the 1-\textit{d} convolution operations with different kernel sizes, formally by:
\begin{equation}
\mathbf{Z}^{i (m)} = \text{Conv1D}(\mathbf{X}^{i}_{P_n}, \mathbf{K}_m), \quad m = 1, \cdots, M,
\end{equation}
where $\mathbf{Z}^{i (m)} \in \mathbb{R}^{T_m \times C}$ represents the learned embeddings at the $m$-th resolution for $i$-th sample where we drop $P_n$ for brevity, $T_m = [T / k_m]$ is the size of transformed temporal embeddings, and $C$ is the number of  channels.

\subsubsection{Multi-resolution Graph Construction}
After obtaining the multi-resolution embeddings, we then aim to construct  multi-resolution graphs for structure learning. Specifically,  we create a series graphs $\{\mathcal{G}^{i (1)}, \cdots, \mathcal{G}^{i (M)}\}$, where each graph $\mathcal{G}^{i (m)} = ({A}^{i(m)}, {X}^{i(m)})$ initialized as a fully-connected graph corresponds to a specific resolution.  ${A}^{i(m)} \in \mathbb{R}^{C*C} $ represents the adjacency matrix, 
and for each graph $\mathcal{G}^{i(m)}$, the node set $\mathcal{V}^{i(m)}$ consists of $C$ nodes, where each node $v^{i(m)}_c \in \mathcal{V}^{i(m)}$ corresponds to a channel in the original series. The node feature matrix ${X}^{i(m)} \in \mathbb{R}^{C \times T_m}$ is formed by the learned embeddings at the $m$-th resolution:
\begin{equation}
{X}^{i(m)} = (\mathbf{Z}^{i (m)})^{\top}.
\end{equation}
The edge set $\mathcal{E}^{i(m)}$ represents the dependencies and correlations among the channels. We initialize the adjacency matrix ${A}^{i(m)} \in \mathbb{R}^{C \times C}$ as a learnable matrix to capture the edge weights between pairs of nodes. The edge weights can be learned adaptively to reflect the dynamic channel correlations at different resolutions.

By constructing multi-resolution graphs, we could obtain a hierarchical representation that captures both channel-level and temporal dynamics at different time scales for the medical time series. This rich representation allows for modeling the complex local and longer contextual spatiotemporal patterns in medical time series.

\vspace{-2mm}
\subsection{Difference Attention Networks for the Baseline Wander in Temporal Dynamics}\label{sec:2}
For the physiological time series in the medical domain, the baseline wander~\cite{MoyYSAA21} - the constant offsets or slow drifts towards baseline measurements - is a common artifact in ECG and EEG recordings. This could make the model capture the less meaningful patterns, i.e., the slow fluctuations around the baseline that may be caused by  accidental factors such as patient movement or respiration. To address this problem, we further propose a new architecture, Difference Attention Networks, to learn the temporal dynamics of medical time series in a more focused fashion. The basic idea is to incorporate the concept of \textit{finite difference} into the attention mechanisms. Note that the difference operates on the temporal dimension  of the input data. 

Specifically, instead of directly applying self-attention to the node features, we compute the differences along the temporal dimension and apply self-attention to these differences.
We first add paddings to the temporal embedding dimensions to get ${X'}^{i(m)} \in \mathbb{R}^{C \times T_m+1}$ obtained from the adaptive graph learning module at the $m$-th resolution, we first compute the first-order finite difference along the temporal dimension:
\begin{equation}
{D}^{i(m)}{(t)} = {X'}^{i(m)}{(t+1)} - {X'}^{i(m)}{(t)}, \quad t = 1, \cdots, T_m,
\end{equation}
where ${D}^{i(m)} \in \mathbb{R}^{C}$ represents the first-order difference of series representation at time step $t$, and ${X'}^{i(m)}{(t)}$ represents the value of ${X'}^{i(m)}$ at $t$.
We then apply the multi-head self-attention mechanism to the difference representations ${D}^{i(m)}$ to learn the temporal dependencies. The self-attention is calculated as:
\begin{equation}
{Attn}^{i(m)} = \text{Softmax}\left(\frac{ {D}^{i(m)} W ^{i(m)}_Q ( {D}^{i(m)} W ^{i(m)}_K )^{\top}}{\sqrt{d}}\right) {D}^{i(m)} W ^{i(m)}_V, 
\end{equation}
where $W^{i(m)}_Q, W^{i(m)}_K$ are learnable weight matrices for queries and keys.
The outputs of difference self-attention are as:
\begin{equation}
 {X}^{i(m)}_{DSA} = \text{DifferenceAttention}({X}^{i(m)}) =  Linear_{DA}( {Attn}^{i(m)}  ),
\end{equation}
where $Linear_{DA}(\cdot)$ represents the final linear layers for output, and $W^{i(m)}_V$ are learnable weight matrices for values in self-attention.
The final outputs of Difference Attention Networks are formed as:
\begin{equation}
    {X}^{i(m)}_{DA} = {X}^{i(m)}_{DSA} + {X'}^{i(m)}.
\end{equation}
The Difference Attention Networks provide a novel way to focus on the temporal changes in medical time series, reducing the influence of slow baseline drifts and highlighting the meaningful patterns. 

\vspace{-2mm}
\subsection{Frequency Convolution Networks for the Multi-view Temporal Representations}\label{sec:3}

Though the difference attention network can help address the baseline wander problem, the processed representations are in the ``difference space'' and it might lose some information in the original data space. To better capture the multi-view information of medical time series signals, we further introduce the frequency convolution networks to enhance the temporal representations~\cite{yi2024frequency,fan2024deep,kunyi_2023survey,fan2022depts}. Note that the temporal convolution networks are operated in parallel with Difference Attention Networks for multi-view information. 

Specifically, given the node feature matrix ${X}^{i(m)} \in \mathbb{R}^{C \times T_m}$ obtained from the multi-resolution graph construction at the $m$-th resolution, we first apply the Fourier transform to convert the temporal signals from the time domain to the frequency domain:
\begin{equation}
\mathcal{X}^{i(m)} = \mathcal{F}({X}^{i(m)}) = \int_{-\infty}^{\infty} {X}^{i(m)}(t) e^{-j2\pi ft} \mathrm{d}t,
\end{equation}
where $\mathcal{F}$ denotes the Fourier transform, $f$ is the frequency variable, $t$ is the integral variable, and $j$ is the imaginary unit. The resulting $\mathcal{X}^{i(m)} \in \mathbb{C}^{C \times S}$ represents the frequency domain representation, where $S$ is the number of frequency components.
Next, we apply Fourier convolution layers to the frequency domain representations to capture the dependencies and patterns in the frequency space:
\begin{equation}
\mathcal{H}^{i(m)} = \text{FourierConvolution}(\mathcal{X}^{i(m)}) = \mathcal{X}^{i(m)} \odot \mathcal{W}^{i(m)},
\end{equation}
where $\mathcal{W}^{i(m)} \in \mathbb{C}^{C \times S}$ represents the learnable convolution kernels in the frequency domain, and $\odot$ denotes the element-wise multiplication.
The resulting $\mathcal{H}^{i(m)} \in \mathbb{C}^{C \times S}$ represents the convolved frequency domain representations.
Finally, we apply the inverse Fourier transform to recover the temporal representations from the frequency domain back to the time domain:
\begin{equation}
{X}^{i(m)}_{FC} = \mathcal{F}^{-1}(\mathcal{H}^{i(m)}) = \int_{-\infty}^{\infty} \mathcal{H}^{i(m)}(f) e^{j2\pi ft} \mathrm{d}f,
\end{equation}
where $\mathcal{F}^{-1}$ denotes the inverse Fourier transform, and ${X}^{i(m)}_{FC} \in \mathbb{R}^{C \times T_m}$ represents the recovered temporal representations.
The frequency convolution networks provide a complementary view of the temporal dynamics by operating in the frequency domain. 

\subsection{Learning Spatial Dynamics with Multi-resolution Graph Transformer}\label{sec:4}


After obtaining the temporal representations from the Difference Attention Networks ${X}^{i(m)}_{DA}$ and the Frequency Convolution Networks ${X}^{i(m)}_{FC}$ at each resolution $m$, we aim to further capture the inter-series (spatial) dependencies based on the explicit structures of multi-resolution graphs. For this aim, we propose the multi-resolution graph transformer model for spatial dependency learning.
Specifically, first, for each resolution, we sum the two learned temporal representations ${X}^{i(m)}_{DA}$ and ${X}^{i(m)}_{FC}$ to obtain the fused multi-view representations at the $m$-th resolution:
\begin{equation}
{X}^{i(m)}_{fused} = {X}^{i(m)}_{DA} + {X}^{i(m)}_{FC}.
\end{equation}

\begin{table*}[ht]
    \centering
    \caption{Results of sample-based evaluation on ADFD dataset (3-classes). The best results are in \textcolor{red}{red} and the second are in \textcolor{blue}{blue}.}
    \vspace{-3mm}
    \scalebox{0.84}{
    \begin{tabular}{c | c |c | c| c |c | c |c |c |c |c }
    \toprule[1.5pt]
       {Models}  &{\textbf{MedGNN}}  &{Medformer} &{iTransformer} &{PatchTST} &{FEDformer}  &{Crossformer} &{FourierGNN} &{CrossGNN}  &{TodyNet} &{SimTSC} \\
    \midrule[1pt]
    {Accuracy} &$\textcolor{red}{\textbf{98.42}}\pm0.04$ &$\textcolor{blue}{\textbf{97.62}}\pm0.17$   &$65.10\pm0.25$  &$66.39\pm0.55$  &$78.77\pm0.91$   &$88.92\pm0.94$      &$67.27\pm0.68$   &$55.38\pm2.44$  &$90.54\pm0.68$  &$90.96\pm0.15$  \\
    
    {Precision} &$\textcolor{red}{\textbf{98.31}}\pm0.02$  &$\textcolor{blue}{\textbf{97.67}}\pm0.13$  &$62.75\pm0.24$  &$65.28\pm0.54$  &$77.75\pm0.85$  &$88.59\pm0.95$   &$65.26\pm0.74$  &$53.02\pm2.74$  &$90.10\pm0.64$  &$90.72\pm0.29$ \\
    
    {Recall} &$\textcolor{red}{\textbf{98.29}}\pm0.05$ &$\textcolor{blue}{\textbf{97.37}}\pm0.24$   &$62.49\pm0.34$   &$65.29\pm0.72$   &$77.97\pm0.89$   &$88.24\pm1.05$     &$65.08\pm0.79$   &$50.41\pm2.91$  &$90.35\pm0.83$  &$90.54\pm0.13$ \\
    
    {F1 score} &$\textcolor{red}{\textbf{98.30}}\pm0.12$ &$\textcolor{blue}{\textbf{97.51}}\pm0.18$  &$62.51\pm0.36$  &$65.15\pm0.54$  &$77.82\pm0.85$  &$88.39\pm1.00$     &$65.08\pm0.77$   &$48.64\pm4.67$   &$90.21\pm0.72$   &$90.61\pm0.16$ \\
    
    {AUROC} &$\textcolor{red}{\textbf{99.93}}\pm0.11$  &$\textcolor{blue}{\textbf{99.84}}\pm0.02$   &$81.63\pm0.20$   &$83.20\pm0.30$   &$92.36\pm0.49$   &$97.40\pm0.35$     &$83.26\pm0.46$  &$70.19\pm2.70$  &$98.04\pm0.23$  &$98.22\pm0.11$  \\
    
    
    \bottomrule[1.5pt]
    \end{tabular}
    }
    \label{tab:sub-dep}
    \vspace{-2mm}
\end{table*}

\begin{table*}[ht]
    \centering
    \caption{Overall results of subject-based evaluation. The best results are in \textcolor{red}{red} and the second best are in \textcolor{blue}{blue}.}
    \vspace{-3mm}
    \scalebox{0.81}{
    \begin{tabular}{l |c |c | c| c | c |c | c |c |c  |c  |c }
    \toprule[1.5pt]
    \multicolumn{2}{c|}{Models}  &{\textbf{MedGNN}}  &{Medformer} &{iTransformer} &{PatchTST} &{FEDformer}  &{Crossformer}  &{FourierGNN} &{CrossGNN} &{TodyNet} &{SimTSC} \\
    \midrule[1pt]
        \multirow{5}*{\rotatebox{90}{ADFD}}  
        &{Accuracy} &$\textcolor{red}{\textbf{56.12}}\pm0.11$  &$\textcolor{blue}{\textbf{53.25}}\pm 1.10$  &$52.13\pm1.37$  &$43.68\pm0.38$  &$46.25\pm0.62$  &$50.42\pm1.26$    &$52.00\pm1.33$  &$47.16\pm1.69$  &$42.34\pm2.48$   &$49.26\pm2.58$  \\
        
        &{Precision} &$\textcolor{red}{\textbf{55.07}}\pm0.09$  &$\textcolor{blue}{\textbf{51.30}}\pm 1.13$   &$46.58\pm1.08$  &$41.60\pm0.91$  &$46.33\pm1.65$   &$45.48\pm2.13$    &$47.20\pm0.87$   &$44.74\pm2.41$ &$42.99\pm2.51$  &$47.04\pm0.05$ \\
        
        &{Recall} &$\textcolor{red}{\textbf{55.47}}\pm0.34$  &$\textcolor{blue}{\textbf{50.76}}\pm0.83$   &$46.96\pm1.02$  &$40.46\pm1.25$  &$44.22\pm1.02$  &$45.80\pm1.82$     &$48.21\pm0.95$    &$41.39\pm1.38$  &$42.48\pm2.57$  &$42.75\pm0.30$  \\
        
        &{F1 score} &$\textcolor{red}{\textbf{55.00}}\pm0.24$  &$\textcolor{blue}{\textbf{50.75}}\pm0.91$  &$46.60 \pm0.90$ &$39.97\pm1.92$  &$43.75\pm0.88$  &$45.46\pm2.03$   &$47.30\pm0.75$    &$39.76\pm1.87$  &$41.05\pm2.03$  &$41.79\pm1.25$  \\
        
        &{AUROC} &$\textcolor{red}{\textbf{74.68}}\pm0.33$ &$\textcolor{blue}{\textbf{70.59}}\pm 1.31$  &$67.14\pm0.97$  &$59.34\pm0.75$  &$62.48\pm1.45$  &$66.24\pm1.86$    &$66.18\pm0.94$    &$58.61\pm1.56$  &$60.69\pm2.63$  &$65.10\pm0.58$  \\
        
    \midrule[1pt]
        \multirow{5}*{\rotatebox{90}{APAVA}}  
        &{Accuracy} &$\textcolor{red}{\textbf{82.60}}\pm 0.35$  &$72.66\pm 6.57 $ &$75.33\pm1.42$  &$68.50\pm1.95$  &$74.87\pm2.07$  &$75.09\pm0.71$  &$68.47\pm7.86$    &$54.40\pm3.15$ &$72.66\pm5.70$  &$\textcolor{blue}{\textbf{82.35}}\pm0.77$ \\
        
        &{Precision} &$\textcolor{red}{\textbf{87.70}}\pm 0.22$  &$73.53\pm 8.23$  &$75.32\pm1.80$  &$77.98\pm1.96$  &$74.45\pm1.45$  &$79.64\pm3.04$    &$67.60\pm8.76$   &$46.37\pm9.06$ &$74.74\pm7.33$  &$\textcolor{blue}{\textbf{85.49}}\pm1.68$  \\
        
        &{Recall} &$\textcolor{red}{\textbf{78.93}}\pm 0.09$  &$70.20\pm 5.46$   &$73.10\pm1.90$   &$61.94\pm2.58$   &$73.24\pm3.32$   &$70.67\pm0.75$      &$66.10\pm7.94$   &$50.60\pm3.19$ &$71.25\pm4.61$  &$\textcolor{blue}{\textbf{78.69}}\pm0.70$ \\
        
        &{F1 score} &$\textcolor{red}{\textbf{80.25}}\pm 0.16$  &$70.69\pm 5.90$ &$73.52\pm1.83$  &$59.25\pm3.79$  &$73.33\pm3.21$   &$71.06\pm0.89$     &$66.36\pm8.13$   &$47.54\pm6.17$ &$70.91\pm5.23$  &$\textcolor{blue}{\textbf{79.90}}\pm0.77$ \\
        
        &{AUROC} &$\textcolor{red}{\textbf{85.93}}\pm 0.26$   &$75.52\pm 7.36$  &$\textcolor{blue}{\textbf{85.79}}\pm1.79$  &$66.32\pm3.20$   &$83.59\pm1.92$     &$79.14\pm4.86$   &$75.03\pm11.62$   &$49.59\pm5.19$  &$78.85\pm4.70$   &$82.37\pm2.79$ \\
        
    \midrule[1pt]
        \multirow{5}*{\rotatebox{90}{PTB}}  
        &{Accuracy} &$\textcolor{red}{\textbf{84.53}}\pm 0.28$  &$79.76\pm1.24$   &$\textcolor{blue}{\textbf{83.41}}\pm2.04$   &$76.33\pm1.06$   &$76.90\pm2.65$   &$81.46\pm3.62$      &$79.70\pm3.30$   &$75.92\pm2.06$ &$77.41\pm3.65$ &$76.99\pm0.98$ \\
        
        &{Precision} &$\textcolor{blue}{\textbf{87.35}}\pm 0.45$  &$81.95\pm0.97$  &$\textcolor{red}{\textbf{87.88}}\pm1.59$   &$79.36\pm1.63$  &$78.94\pm3.57$ &$84.99\pm3.56$    &$81.46\pm4.58$ &$78.35\pm1.79$ &$82.50\pm2.10$  &$80.61\pm1.70$ \\
        
        &{Recall} &$\textcolor{red}{\textbf{77.90}}\pm 0.66$  &$71.50\pm2.08$   &$\textcolor{blue}{\textbf{75.67}}\pm3.06$  &$66.05\pm1.52$   &$67.30\pm 3.86$   &$73.32\pm5.41$    &$71.53\pm4.55$  &$65.60\pm3.22$ &$67.12\pm6.09$  &$66.92\pm1.82$ \\
        
        &{F1 score} &$\textcolor{red}{\textbf{80.40}}\pm 0.62$  &$73.44\pm2.26$   &$\textcolor{blue}{\textbf{78.22}}\pm3.29$   &$67.16\pm1.89$   &$68.52\pm 4.56$   &$75.34\pm5.84$     &$73.33\pm5.19$  &$66.49\pm3.81$  &$67.81\pm7.62$  &$68.19\pm2.21$ \\
        
        &{AUROC}  &$\textcolor{red}{\textbf{93.31}}\pm 0.46$  &$\textcolor{blue}{\textbf{93.13}}\pm0.50$   &$90.61\pm2.46$   &$88.48\pm0.85$   &$86.39\pm 2.86$  &$89.71\pm3.71$    &$84.93\pm3.44$   &$89.93\pm 0.47$ &$91.72\pm1.55$  &$88.60\pm0.66$ \\
        
    \midrule[1pt]
        \multirow{5}*{\rotatebox{90}{PTB-XL}}  
        &{Accuracy} &$\textcolor{red}{\textbf{73.87}}\pm 0.18$ &$72.87\pm0.23$   &$69.13\pm0.21 $  &$73.15\pm0.24$   &$57.23\pm 9.48$   &$\textcolor{blue}{\textbf{73.23}}\pm0.18$   &$63.49\pm 0.88$   &$63.09\pm 0.63$ &$72.21\pm0.65$  &$71.87\pm0.91$ \\
        
        &{Precision} &$\textcolor{red}{\textbf{66.26}}\pm 0.29$   &$64.24\pm0.42$   &$59.46\pm0.44$   &$\textcolor{blue}{\textbf{65.47}}\pm0.50$   &$52.32\pm 6.38$   &$64.92\pm0.58$    &$52.05\pm1.21 $  &$49.60\pm 0.93$  &$64.30\pm0.68$   &$63.61\pm1.73$  \\
        
        &{Recall} &$\textcolor{red}{\textbf{61.13}} \pm 0.23$   &$60.09\pm0.30$   &$54.56\pm0.23$   &$\textcolor{blue}{\textbf{60.64}}\pm0.44$   &$48.76\pm7.21$   &$\textcolor{red}{\textbf{61.13}}\pm0.45$      &$47.85\pm1.13$    &$44.94\pm0.65$  &$57.12\pm1.66$  &$58.86\pm0.99$  \\
        
        &{F1 score} &$\textcolor{red}{\textbf{62.54}}\pm 0.20$   &$61.70\pm0.23$   &$56.24\pm0.24$   &$62.46\pm0.28$   &$47.75\pm8.36$   &$\textcolor{blue}{\textbf{62.51}}\pm0.31$     &$48.92\pm1.19 $  &$45.08\pm0.44$  &$58.94\pm1.75$  &$60.19\pm0.60$ \\
        
        &{AUROC} &$\textcolor{red}{\textbf{90.21}} \pm 0.15$   &$89.64\pm0.19$   &$86.67\pm0.18$   &$89.61\pm0.15$   &$82.15\pm4.25$   &$\textcolor{blue}{\textbf{89.96}}\pm0.16$    &$82.45\pm0.86$   &$81.43\pm0.33$ &$89.14\pm0.44$  &$88.19\pm0.78$ \\
        
    \midrule[1pt]
        \multirow{5}*{\rotatebox{90}{TDBRAIN}}  
        &{Accuracy} &$\textcolor{red}{\textbf{91.04}} \pm 0.09$   &$88.71\pm1.26$   &$74.87\pm1.30$   &$73.90\pm4.40$   &$77.56\pm1.67$   &$82.48\pm1.41$      &$76.04\pm3.14$  &$70.60\pm3.31$  &$\textcolor{blue}{\textbf{89.58}}\pm2.15$  &$89.06\pm1.67$  \\
        
        &{Precision} &$\textcolor{blue}{\textbf{91.15}}\pm 0.12$  &$88.84\pm1.13$   &$74.99\pm1.33$  &$74.21\pm4.26$   &$78.00\pm1.84$   &$82.71\pm1.21$     &$76.22\pm3.22$   &$70.71\pm3.21$ &$\textcolor{red}{\textbf{90.21}}\pm1.39$  &$89.41\pm0.86$ \\
        
        &{Recall} &$\textcolor{red}{\textbf{91.04}} \pm 0.20$   &$88.71\pm1.26$   &$74.87\pm1.30$  &$73.90\pm4.40$   &$77.56\pm1.67$   &$82.48\pm1.41$   &$76.04\pm3.14$   &$70.60\pm3.31$ & $\textcolor{blue}{\textbf{89.58}}\pm3.87$  &$89.06\pm0.50$  \\
        
        &{F1 score} &$\textcolor{red}{\textbf{91.04}} \pm 0.08$   &$88.70\pm1.27$   &$74.85\pm1.29$   &$73.79\pm4.48$   &$77.48\pm1.65$   &$82.44\pm1.44$  &$76.00\pm3.14$     &$70.55\pm3.38$  &$\textcolor{blue}{\textbf{89.54}}\pm4.69$   &$89.04\pm1.01$  \\
        
        &{AUROC} &$96.74 \pm 0.04$   &$96.24\pm0.59$   &$83.66\pm1.14$   &$80.93\pm6.12$   &$86.41\pm1.45$   &$91.58\pm0.88$    &$84.11\pm2.83$   &$78.42\pm3.43$ &$\textcolor{red}{\textbf{97.41}}\pm1.00$ &$\textcolor{blue}{\textbf{97.28}}\pm1.69$  \\
        
    \bottomrule[1.5pt]
    \end{tabular}
    }
    \label{tab:sub-indep}
    \vspace{-4mm}
\end{table*}

Next, inspired by Graph Transformer~\cite{gt_survey}, we apply the local attention mechanisms first and then pass the representations to the graph neural networks for processing. Specifically,
the local attention for graph processing is first computed as:
\begin{equation}
    \alpha_{pq} = \frac{\exp(g(\mathbf{x}_p, \mathbf{x}_q) \cdot b_{pq})}{\sum_{k \in \mathcal{N}(v_p)}\exp(g(\mathbf{x}_p, \mathbf{x}_k) \cdot b_{pk})},
\end{equation}
where $\alpha_{pq}$ is the attention score, $x_p$ and $x_q$ are their node features, $g$ is a function that computes the similarity between two nodes, i.e., the dot-product. $b_{pq}$ is a local attention bias term. Note that the neighborhood is dynamically learned in Section \ref{sec:1}. Then we can calculate the graph representations by:
\begin{equation}
{X}^{i(m)}_{GA} = \text{LocalAttention}({X}^{i(m)}_{fused}) = \mathbf{\alpha} {X}^{i(m)}_{fused},
\end{equation}
where ${X}^{i(m)}_{GA} \in \mathbb{R}^{C \times T_m}$ represents the spatially attended representations. The multi-head self-attention allows the model to attend to different spatial locations and capture the dependencies among different time series channels.
After obtaining the spatially attended representations, we apply graph convolution networks to further incorporate the graph structure information at each resolution. Given the constructed graph $\mathcal{G}^{i(m)} = ({A}^{i(m)}, {X}^{i(m)}_{GA})$ at the $m$-th resolution, we perform graph convolution as follows:
\begin{equation}
{X}^{i(m)}_{GT} = \text{GraphConv}(\hat{{A}}^{i(m)},  {X}^{i(m)}_{GA}, {W}^{i(m)}_{GT}),
\end{equation}
where $\hat{{A}}^{i(m)} = \tilde{{D}}^{-\frac{1}{2}} \tilde{{A}}^{i(m)} \tilde{{D}}^{-\frac{1}{2}}$ is the normalized adjacency matrix, $\tilde{{A}}^{i(m)} = {A}^{i(m)} + {I}$ is the adjacency matrix with self-loops, ${I}$ is the identity matrix, $\tilde{{D}}$ is the diagonal degree matrix of $\tilde{{A}}^{i(m)}$, ${W}^{i(m)}_{GT} \in \mathbb{R}^{T_m \times T_m}$ is the learnable weight matrix for graph convolution.
The resulting ${X}^{i(m)}_{GT} \in \mathbb{R}^{C \times T_m}$ represents the graph convoluted representations at the $m$-th resolution.
Since we have multiple resolutions, we need to fuse the representations obtained from different resolutions to capture the multi-resolution dynamics. We achieve this by applying average pooling  across the resolution dimension:
\begin{equation}
{X}^{i}_{MRFused} = \text{AvgPool}({{X}^{i(1)}_{GT}, \ldots, {X}^{i(M)}_{GT}}),
\end{equation}
where $\text{AvgPool}(\cdot)$ denotes the average pooling operation, and the output ${X}^{i}_{MRFused} \in \mathbb{R}^{C \times T}$ represents the fused multi-resolution representations.
Finally, we feed the fused multi-resolution representations ${X}^{i}_{MRFused}$ into a linear or fully-connected layer followed by a softmax activation function for classification:
\begin{equation}
\hat{{y}}^{i} = \text{Softmax}(\text{Linear}_{GT}({X}^{i}_{MRFused})),
\end{equation}
where $\text{Linear}_{GT}(\cdot)$ is a learnable linear layer that maps the fused representations to the output space, and $\hat{{y}}^{i} \in \mathbb{R}^{K}$ represents the predicted probabilities for the $K$ classes.
\vspace{-1.5mm}
\section{Experiments}
In this section, we perform extensively experiments with five real-world medical time series benchmarks to assess the performance of our proposed MedGNN.
Furthermore, we conduct thorough analytical experiments and visualization studies concerning the different components of the MedGNN framework.

\vspace{-2mm}
\subsection{Experimental Settings}

\subsubsection{Datasets}
We conduct empirical analyses on five representative medical datasets,
i.e., ADFD~\cite{miltiadous2023dataset}, APAVA~\cite{escudero2006analysis}, TDBRAIN~\cite{van2022two}, PTB~\cite{physiobank2000physionet}, and PTB-XL~\cite{wagner2020ptb}. 
These datasets include three EEG datasets and two ECG datasets.
The data preprocessing and train-validation-test split are following the previous work~\cite{wang2024medformer}.
For further details on the datasets, please refer to the Appendix~\ref{appendix_dataset}.

\subsubsection{Baselines}
We compare our proposed MedGNN with the representative and state-of-the-art models for time series classification. 
We choose the baseline methods from two categories: (1) GNN-based models, which include TodyNet~\cite{liu2024todynet}, SimTSC~\cite{zha2022towards}, FourierGNN~\cite{yi2024fouriergnn}, and CrossGNN~\cite{huang2023crossgnn}; (2) Transformer-based models, including iTransformer~\cite{liu2023itransformer}, PatchTST~\cite{nie2022time}, FEDformer~\cite{zhou2022fedformer}, Crossformer~\cite{zhang2022crossformer}, Autoformer~\cite{wu2021autoformer} and more recent Medformer~\cite{wang2024medformer}. Further details about the baselines can be found in Appendix \ref{appendix_baseline}.

\subsubsection{Implementation Details}
All experiments are implemented using Pytorch 2.2~\cite{paszke2019pytorch} and conducted on 4 GeForce RTX 4090 GPUs. We employ cross-entropy loss as the loss function and present five metrics: accuracy, precision, recall, F1 score, and AUROC. Further implementation details are presented in Appendix \ref{appendix_evaluation_metrics} and \ref{appendix_implementation}.

\begin{table*}[!h]
    \centering
    \caption{Ablation studies on the effects of Difference Attention (DA) under the subject-based setup.}
    \vspace{-3mm}
    \scalebox{0.83}{
    \begin{tabular}{l  c |c c c c| c c c c |c c c c }
    \toprule[1.5pt]
       \multicolumn{2}{c|}{Datasets}  &\multicolumn{4}{c|}{APAVA}  &\multicolumn{4}{c|}{TDBRAIN} &\multicolumn{4}{c}{PTB-XL}\\
       \cmidrule(r){1-2}\cmidrule(r){3-6}\cmidrule(r){7-10}\cmidrule(r){11-14}
       \multicolumn{2}{c|}{Metrics} &Accuracy &Precision &Recall &F1 score &Accuracy &Precision &Recall & F1 score &Accuracy &Precision &Recall & F1 score \\
    \midrule[1pt]
    \multicolumn{2}{c|}{w DA} &82.60   &87.70  &78.93  &80.25  &91.04   &91.15   &91.04   &91.04   &73.87  &66.26   &61.13 &62.54    \\
    \multicolumn{2}{c|}{w/o DA} &69.88  &69.94  &66.20  &66.38    &70.52   &70.86   &70.52  &70.40   &69.33   &60.05   & 56.07 &57.63  \\
    \midrule[1pt]
    
    \multicolumn{2}{c|}{\textcolor{red}{\textbf{Improvement}}} &\textcolor{red}{\textbf{18.20\%}}  &\textcolor{red}{\textbf{25.40\%}}  &\textcolor{red}{\textbf{19.24\%}}  &\textcolor{red}{\textbf{20.89\%}}  &\textcolor{red}{\textbf{29.10\%}}   &\textcolor{red}{\textbf{28.63\%}}   &\textcolor{red}{\textbf{29.10\%}} &\textcolor{red}{\textbf{29.31\%}}  &\textcolor{red}{\textbf{6.54\%}}   &\textcolor{red}{\textbf{10.35\%}}  &\textcolor{red}{\textbf{9.02\%}}  &\textcolor{red}{\textbf{8.52\%}}  \\
    
    \bottomrule[1.5pt]
    \end{tabular}
    }
    \label{tab:abl_differential}
    \vspace{-3mm}
\end{table*}

\vspace{-1mm}
\subsection{Main Results}
We present results of our proposed MedGNN compared to several representative baselines with two different experimental setup (e.g., sample-based and subject-based evaluation) in Tables~\ref{tab:sub-dep} and \ref{tab:sub-indep}.
Note that the percentage symbol (\%) is omitted in the experimental results, and this will not be repeated below.

\subsubsection{Sample-based Evaluation}
In this setup, the training, validation, and test sets are divided based on test samples. Samples from different subjects are randomly shuffled and then assigned to the corresponding sets. It is a reasonable setting in real-world medical scenarios because many patients return for multiple visits even though it might have the information leakage problem from the perspective deep learning. Since this setting is easier than the subject-based evaluation, we only assess the ADFD dataset to enable a simple comparison for reference. Table \ref{tab:sub-dep} has shown the overall results sample-based evaluation of MedGNN compared with several baseline algorithms. We can easily observe that our proposed MedGNN outperforms all the baselines as illustrated in and achieves the best performance in five different metrics. 


\vspace{-1mm}
\subsubsection{Subject-based Evaluation} The training, validation, and test sets are split based on subjects (patients) under this setup. Each subject, along with their corresponding samples, is assigned to one of the three sets (training, validation, or test) according to a predefined ratio or subject IDs. This ensures that samples from the same subject only appear in one set, preventing overlap between the sets.
Table \ref{tab:sub-indep} presents the results on five datasets under the subject-based setup. Overall, our model achieves leading perfomance on most datasets, achieving 22 top-1 and 2 top-2 out of 25 in total across five datasets. 
Notably, MedGNN ranks first in terms of F1 score across all five datasets, demonstrating that MedGNN is not only accurate but also achieves a good balance between precision and recall, indicating its exceptional robustness in handling classification tasks.
This balance is particularly important in critical applications such as medical diagnosis, where both false positives (misclassifying healthy individuals as sick) and false negatives (misclassifying sick individuals as healthy) can have serious consequences.

\begin{figure}[!t]
    \vspace{-2mm}
    \centering
    \subfigure[ADFD-Subject]
    {
        \centering
        \includegraphics[width=0.46\linewidth]{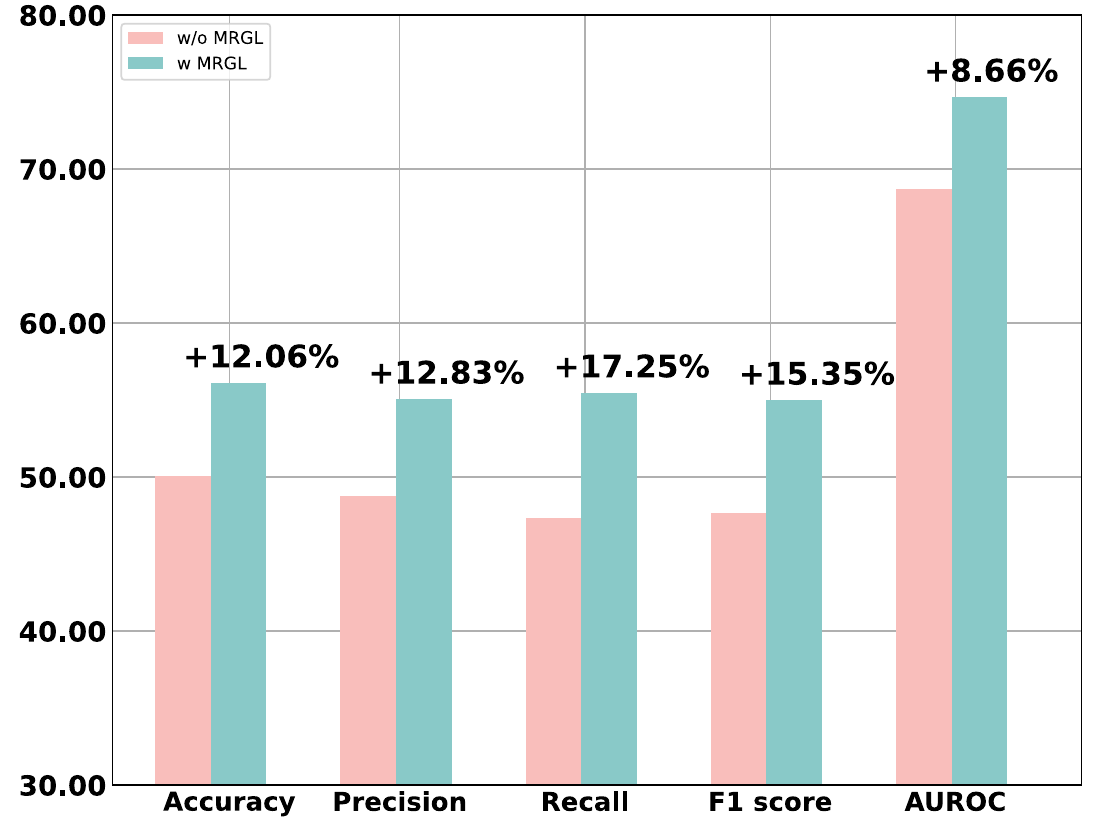}
    }
    \subfigure[APAVA-Subject]
    {
        \centering
        \includegraphics[width=0.46\linewidth]{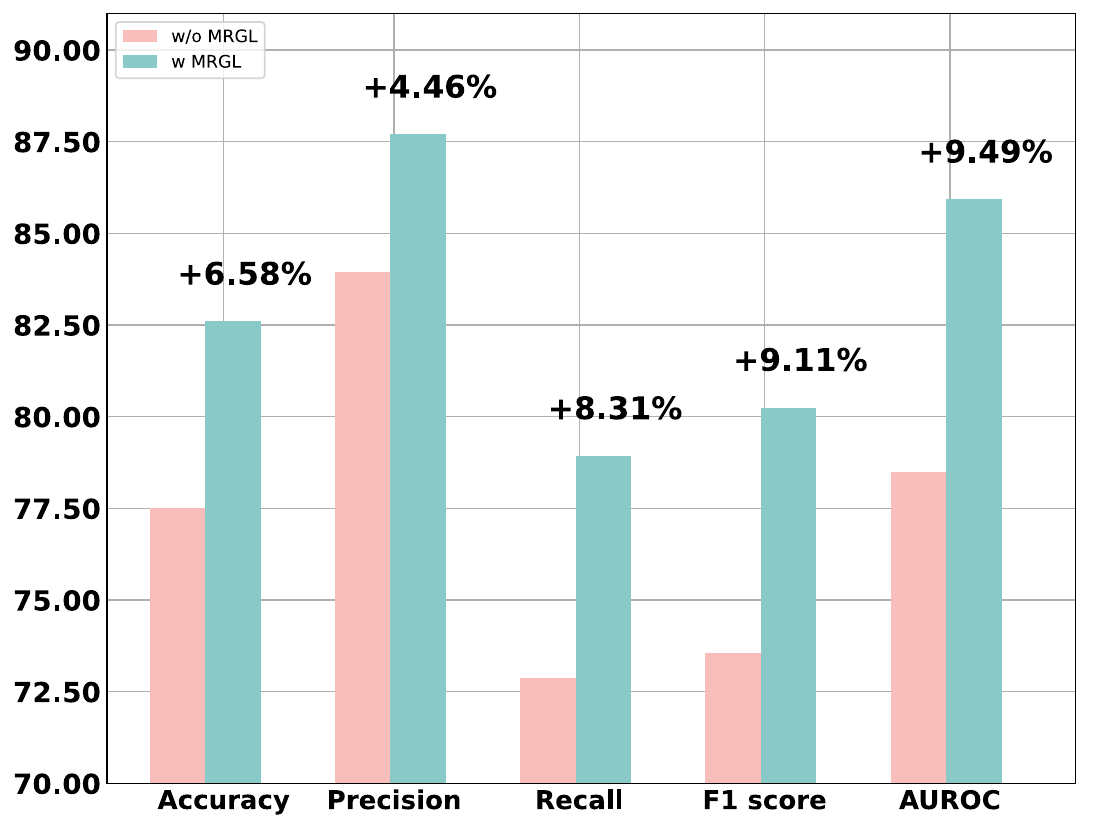}
    }
    \vspace{-5mm}
    \caption{Ablation study of Multi-Resolution Graph Learning (MRGL) under the Subject-based setup.}
    \label{fig:abl_GNN}
    \vspace{-5mm}
\end{figure}

\vspace{-2mm}
\subsection{Ablation Studies}

\subsubsection{Study of the Multi-Resolution Graph Learning}
We explore the impact of multi-resolution graph learning (MRGL) on MedGNN's performance across various metrics. To assess its effectiveness, we compare two versions of the model: one with MRGL enabled and one without it. 
The MRGL-enhanced version demonstrates substantial performance gains as shown in Figure \ref{fig:abl_GNN}, particularly in recall and F1 score, indicating the model's improved ability to detect and balance positive predictions with accurate classifications. This improvement highlights MRGL's capacity to capture multi-resolution dependencies effectively, improving the model's robustness in handling complex medical time series.

\subsubsection{Study of the Frequency Convolution Networks.} 
\begin{figure}[!t]
    \vspace{-4mm}
    \centering
    \subfigure[TDBRAIN-Subject]
    {
        \centering
        \includegraphics[width=0.32\linewidth]{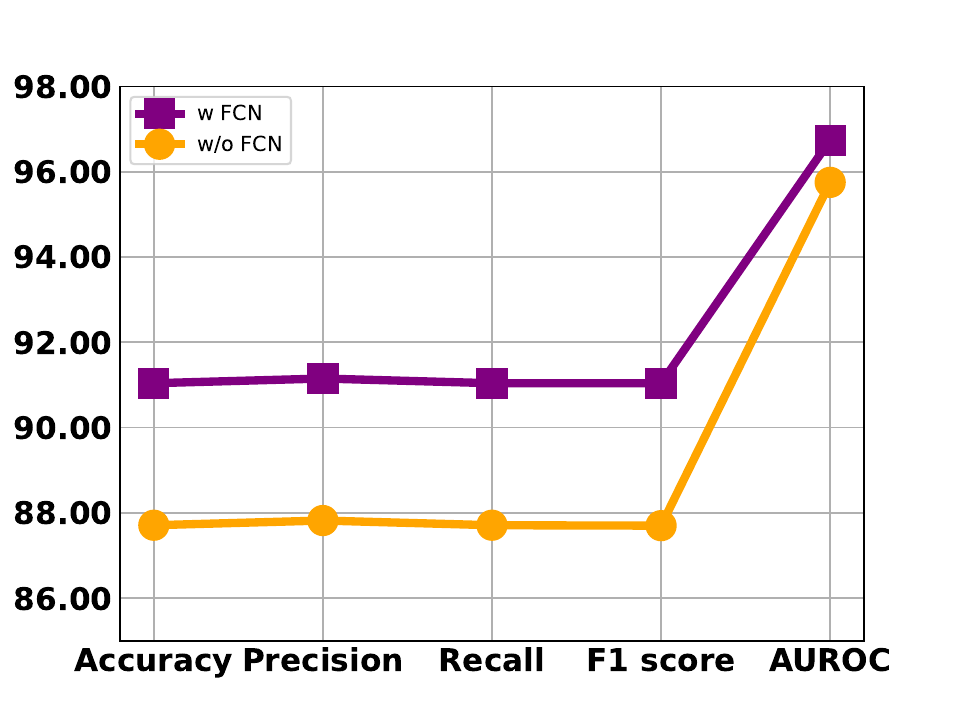}
    }
    \hspace{-4mm}
    \subfigure[ADFD-Subject]
    {
        \centering
        \includegraphics[width=0.32\linewidth]{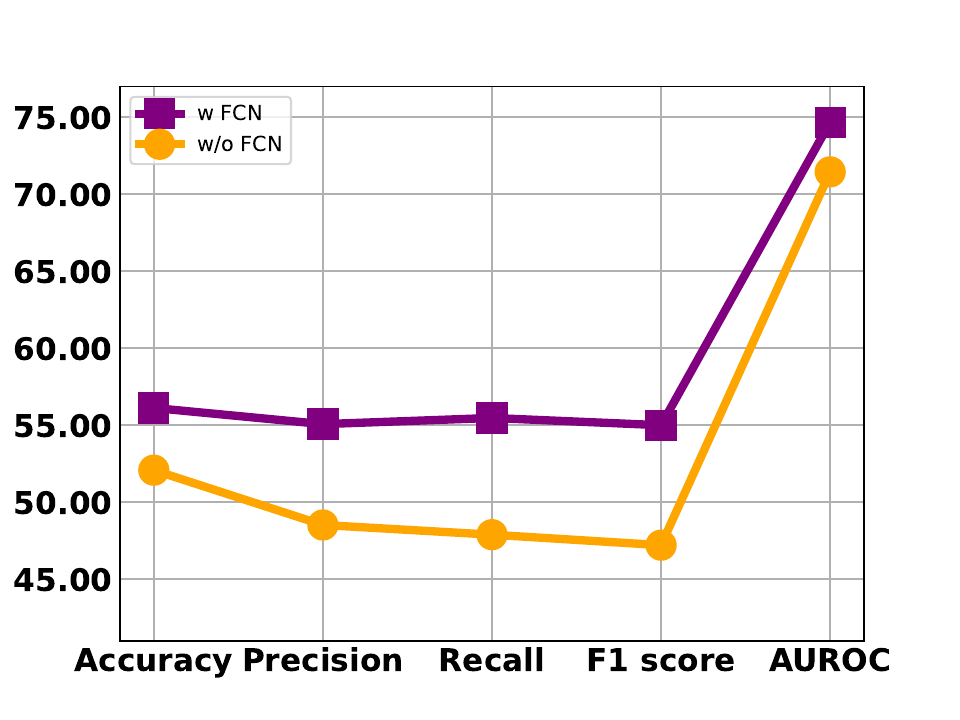}
    }
    \hspace{-4mm}
    \subfigure[PTB-Subject]
    {
        \centering
        \includegraphics[width=0.32\linewidth]{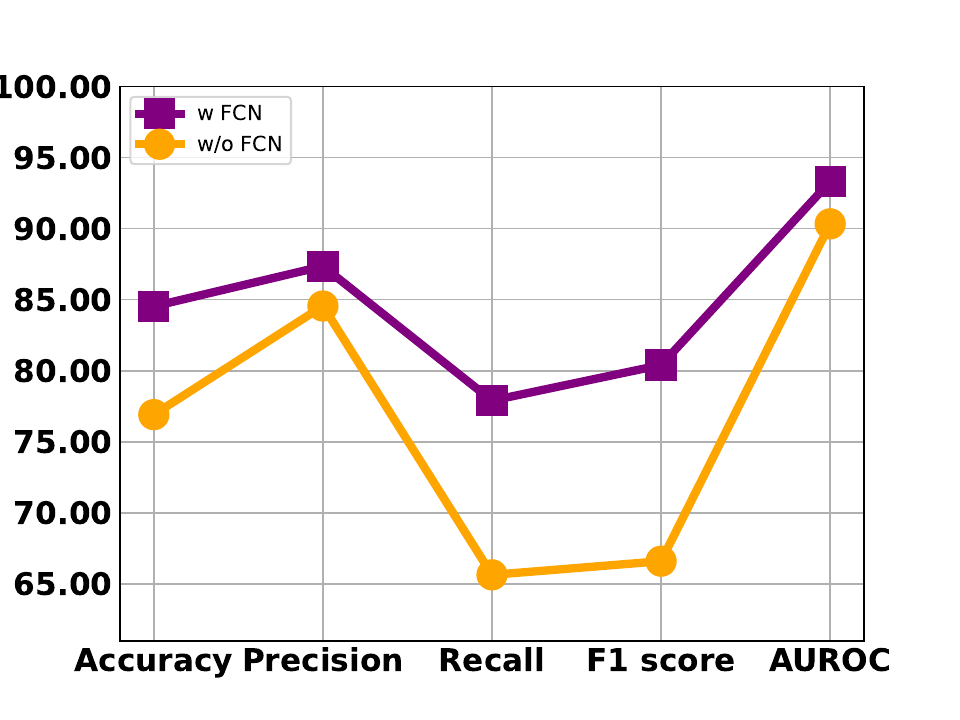}
    }
    \vspace{-4mm}
    \caption{Ablation study of Frequency Convolution Networks (FCN) under the Subject-based setup.}
    \label{fig:abl_FL}
    \vspace{-4mm}
\end{figure}

To evaluate the impact of the Frequency Convolution Networks (FCN) within MedGNN, we perform ablation experiments, as shown in Figure \ref{fig:abl_FL}, where "w/o FCN" denotes the version without FCN. The results clearly demonstrate that incorporating FCN improves MedGNN’s performance across all metrics on the experimental datasets. Moreover, frequency convolution helps prevent potential information loss from relying solely on difference attention, further enhancing the temporal representation.

\subsubsection{Study of the Difference Attention Networks.} In this section, we aim to investigate the effectiveness of the difference attention in the MedGNN framework. Table \ref{tab:abl_differential} has shown the performance comparison of the variant on two EEG datasets (APAVA, TDBRAIN) and one ECG dataset. The version equipped with difference attention achieves improvements of 23.65\%, 27.02\%, 24.17\%, and 25.10\% in the metrics of Accuracy, Precision, Recall, and F1 score on EEG datasets in average, while 6.54\%, 10.35\%, 9.02\% , and 8.52\% respectively on ECG. 
This indicates that difference attention can enhance the model's performance in medical time series classification tasks by minimizing the impact of slow baseline drifts and bringing important patterns to the forefront. 


\begin{figure}[!h]
    \vspace{-3mm}
    \centering
    \subfigure[APAVA-Subject]
    {
        \centering
        \includegraphics[width=0.45\linewidth]{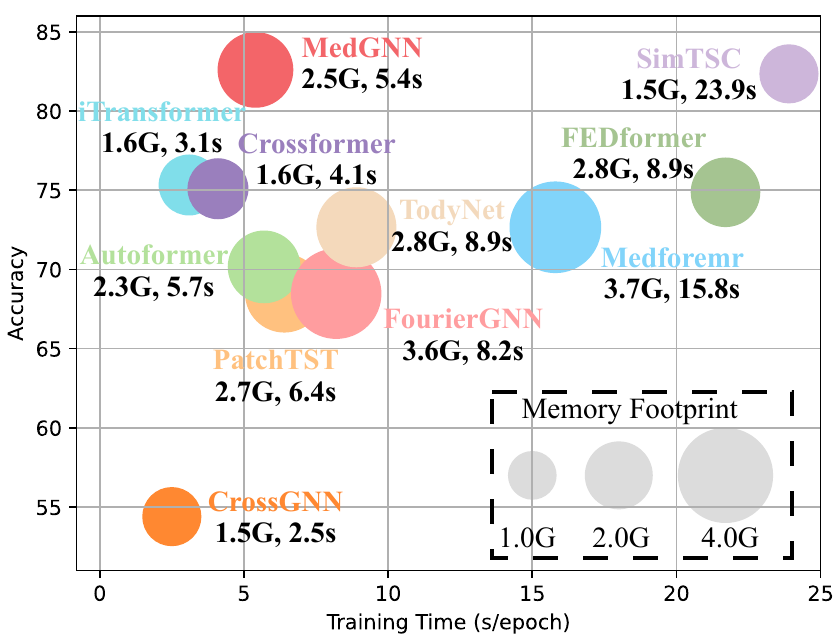}
    }
    \subfigure[TDBRAIN-Subject]
    {
        \centering
        \includegraphics[width=0.45\linewidth]{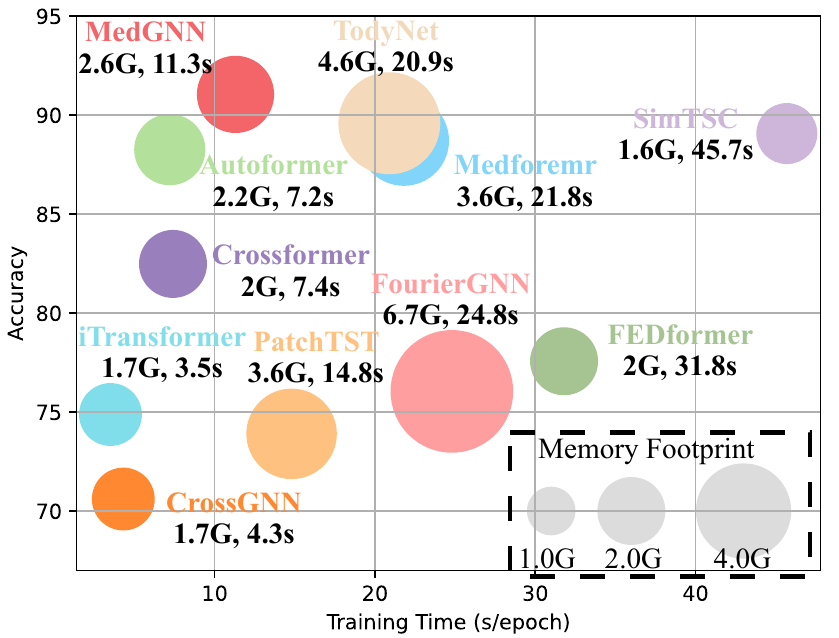}
    }
    \vspace{-4mm}
    \caption{Model effectiveness and efficiency comparison on two datasets under the Subject-based setup.}
    \label{fig:efficiency}
    \vspace{-4mm}
\end{figure}

\vspace{-1mm}
\subsection{Additional Experiments}

\begin{figure*}[!h]
    \vspace{-2mm}
    \centering
    \subfigure[resolution=2]
    {
        \centering
        \includegraphics[width=0.235\linewidth]{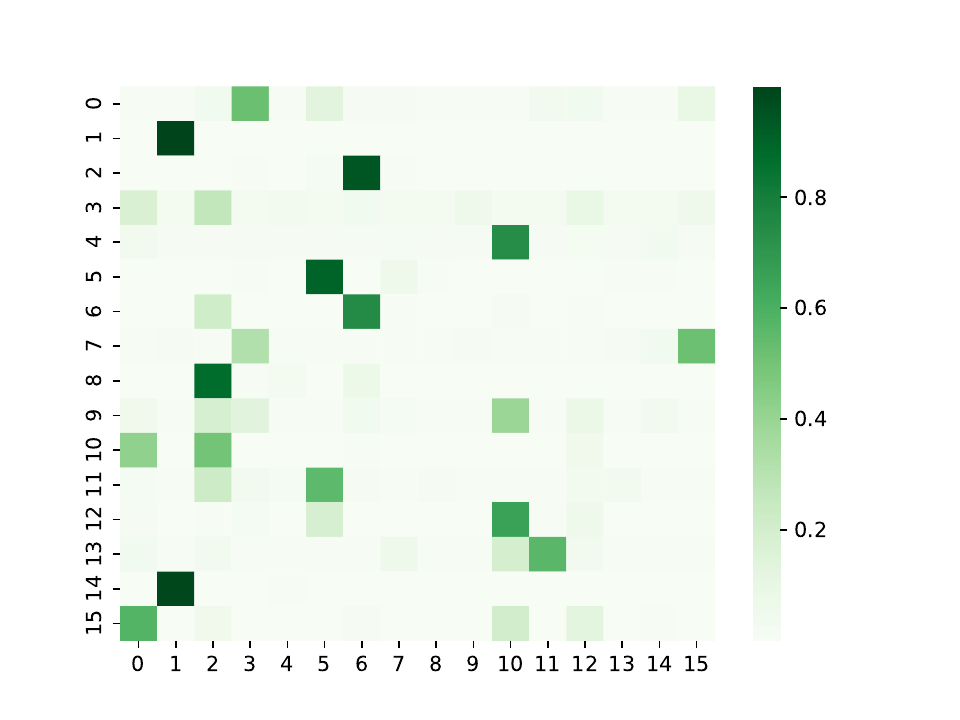}
    }
    \subfigure[resolution=4]
    {
        \centering
        \includegraphics[width=0.235\linewidth]{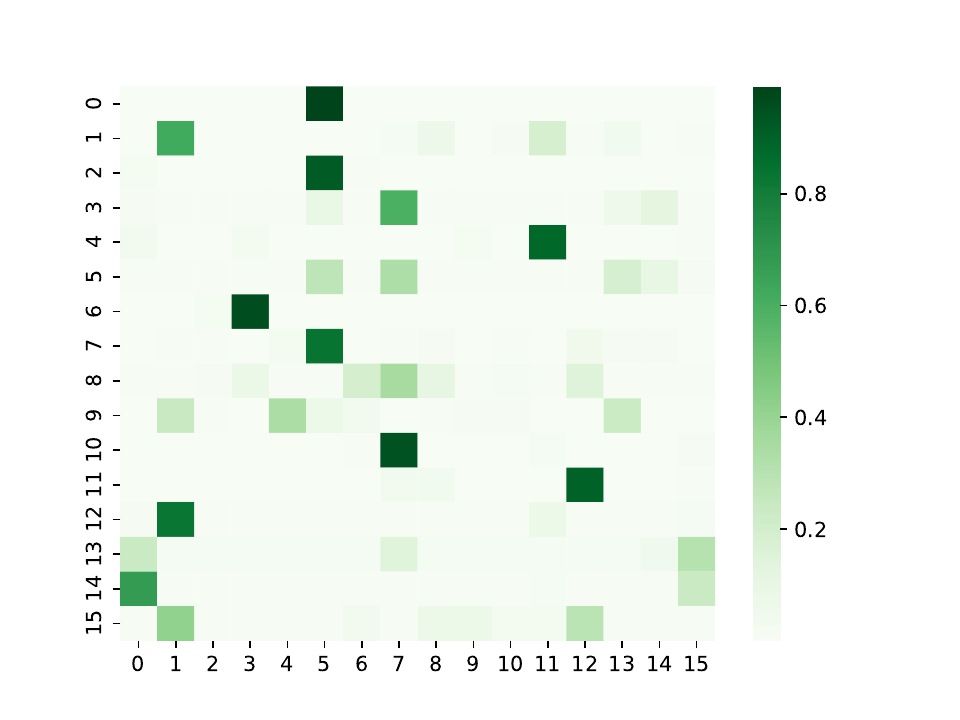}
    }
    \subfigure[resolution=6]
    {
        \centering
        \includegraphics[width=0.235\linewidth]{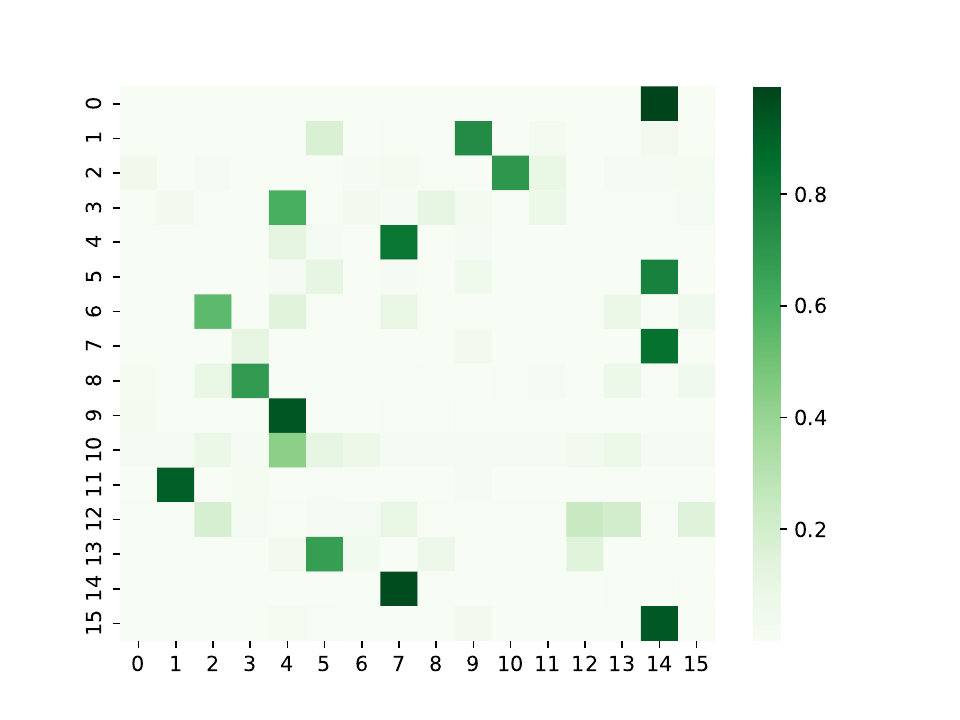}
    }
    \subfigure[resolution=8]
    {
        \centering
        \includegraphics[width=0.235\linewidth]{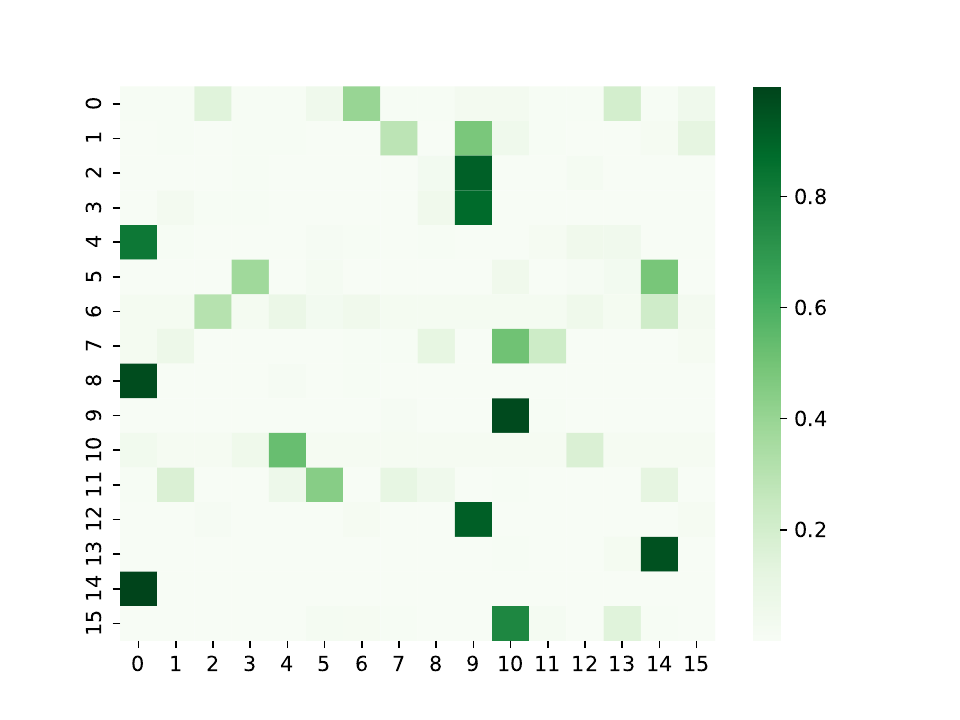}
    }
    \vspace{-4.5mm}
    \caption{Adjacent matrices of multi-resolution graphs learned from APAVA dataset.}
    \label{fig:Adj_APAVA}
    \vspace{-3mm}
\end{figure*}

\begin{figure*}[!h]
    \vspace{-3mm}
    \centering
    \subfigure[resolution=2]
    {
        \centering
        \includegraphics[width=0.187\linewidth]{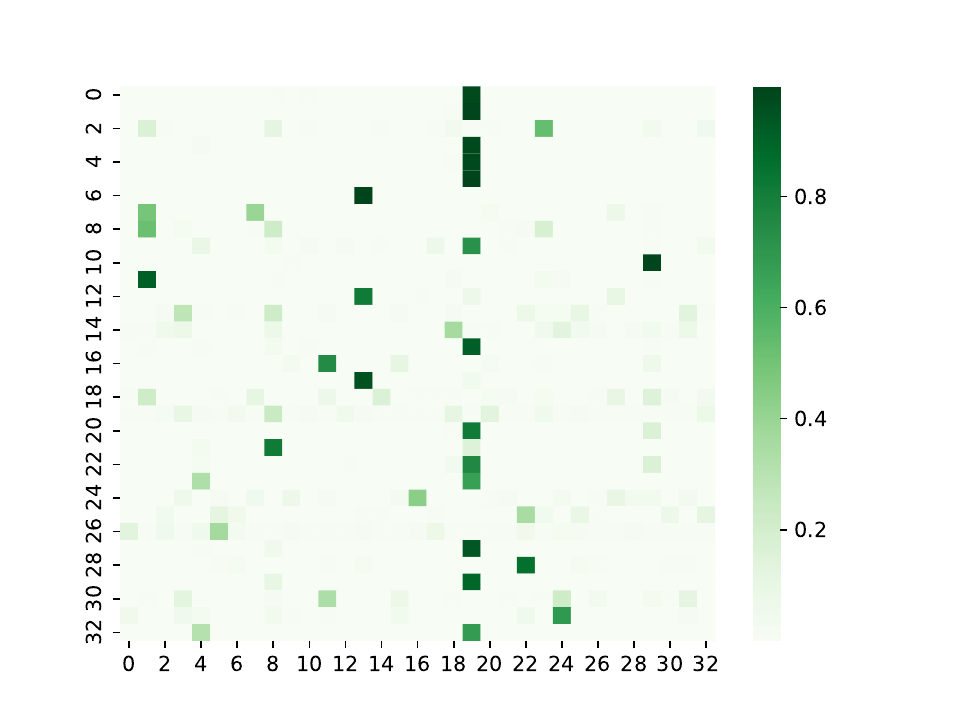}
    }
    \subfigure[resolution=4]
    {
        \centering
        \includegraphics[width=0.187\linewidth]{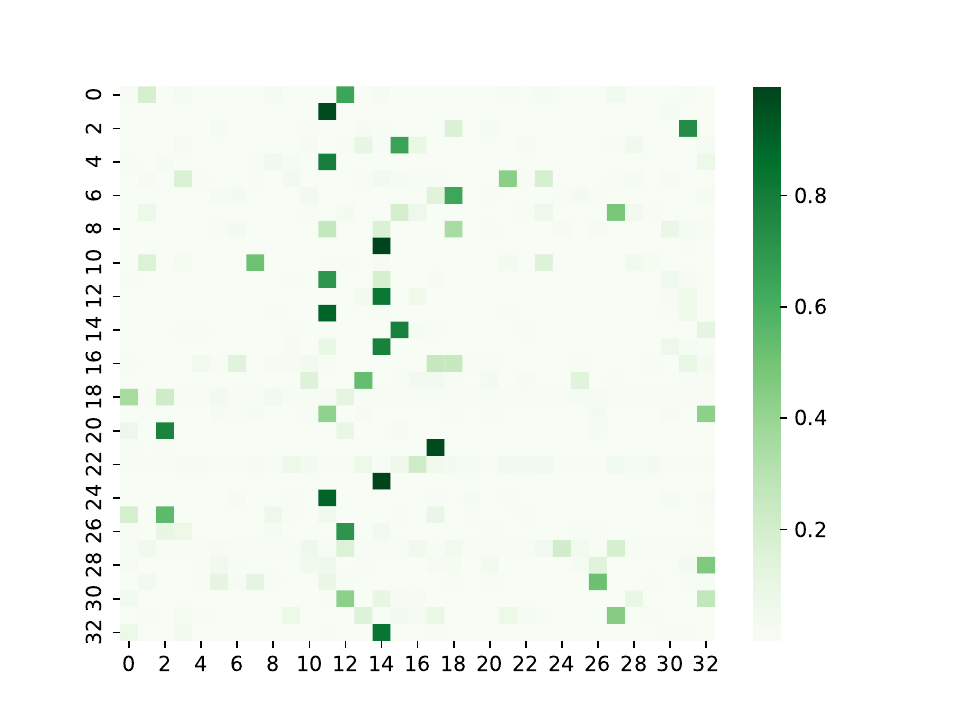}
    }
    \subfigure[resolution=6]
    {
        \centering
        \includegraphics[width=0.187\linewidth]{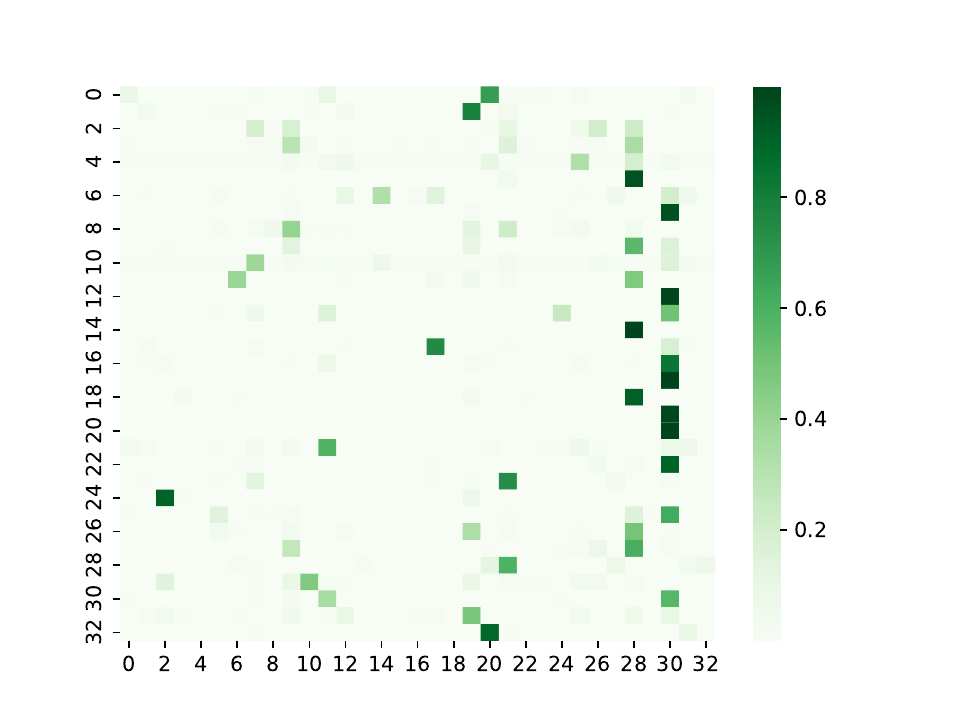}
    }
    \subfigure[resolution=8]
    {
        \centering
        \includegraphics[width=0.187\linewidth]{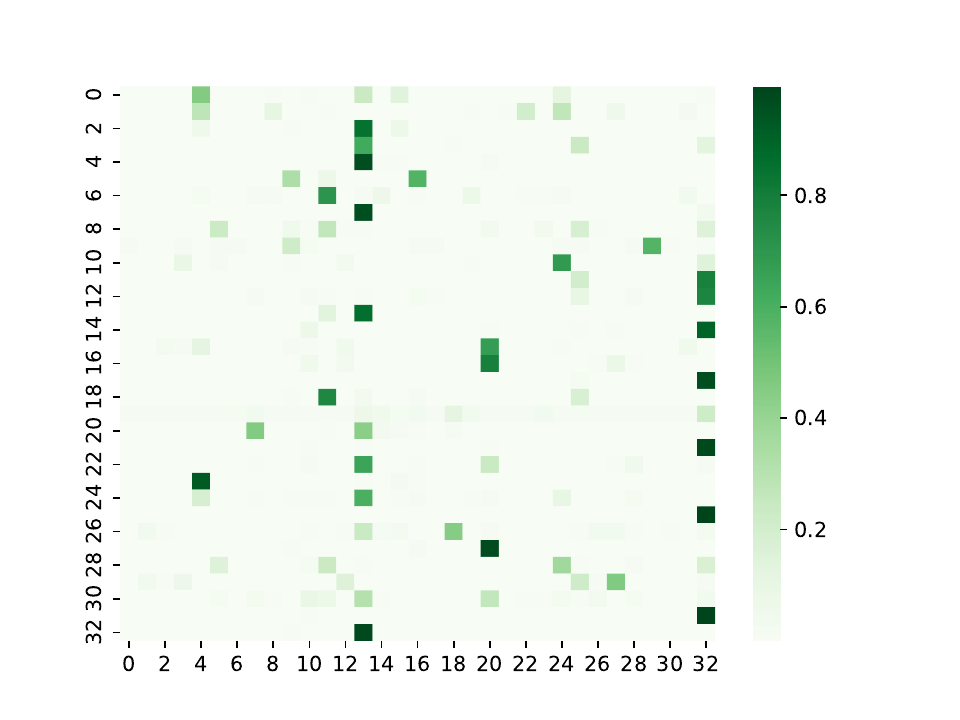}
    }
    \subfigure[resolution=10]
    {
        \centering
        \includegraphics[width=0.187\linewidth]{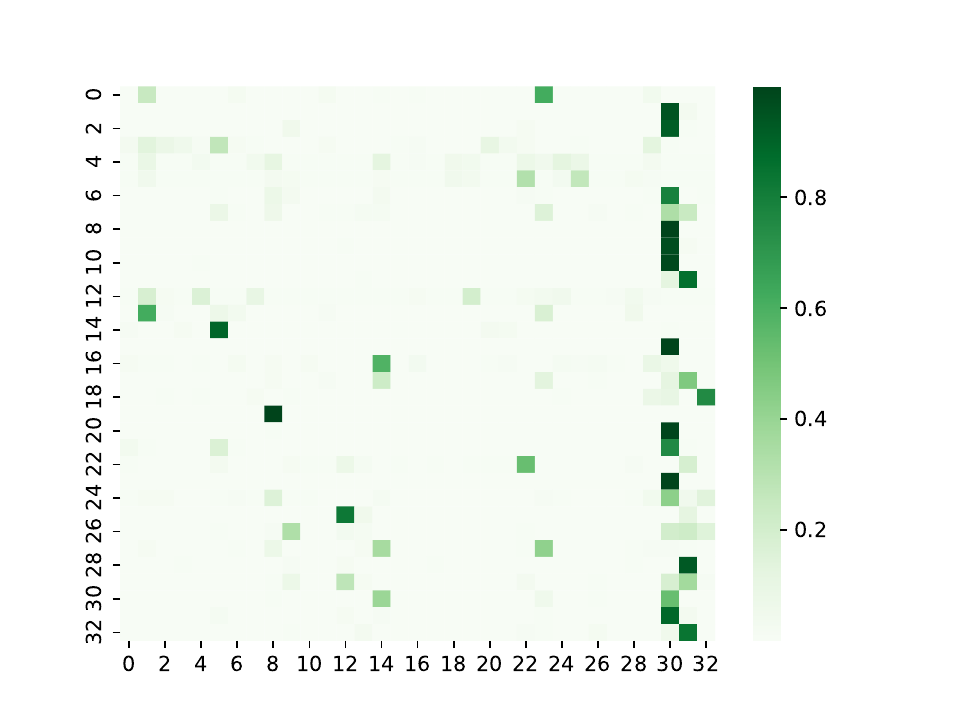}
    }
    \vspace{-4.5mm}
    \caption{Adjacent matrices of multi-resolution graphs learned from TDBRAIN dataset.}
    \label{fig:Adj_TDBRAIN}
    \vspace{-3mm}
\end{figure*}

\begin{figure}[!h]
    \vspace{-3mm}
    \centering
    \subfigure[APAVA-Subject]
    {
        \centering
        \includegraphics[width=0.47\linewidth]{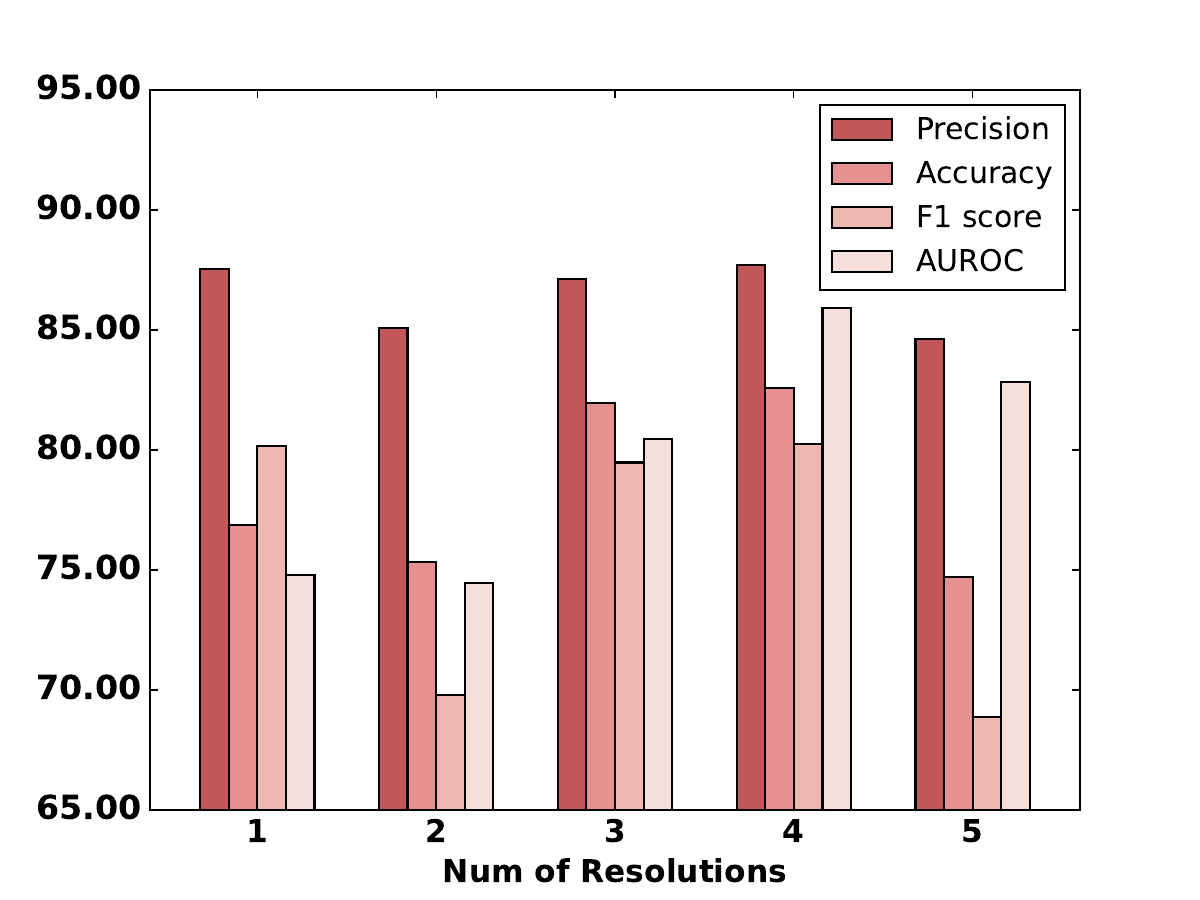}
    }
    \subfigure[TDBRAIN-Subject]
    {
        \centering
        \includegraphics[width=0.47\linewidth]{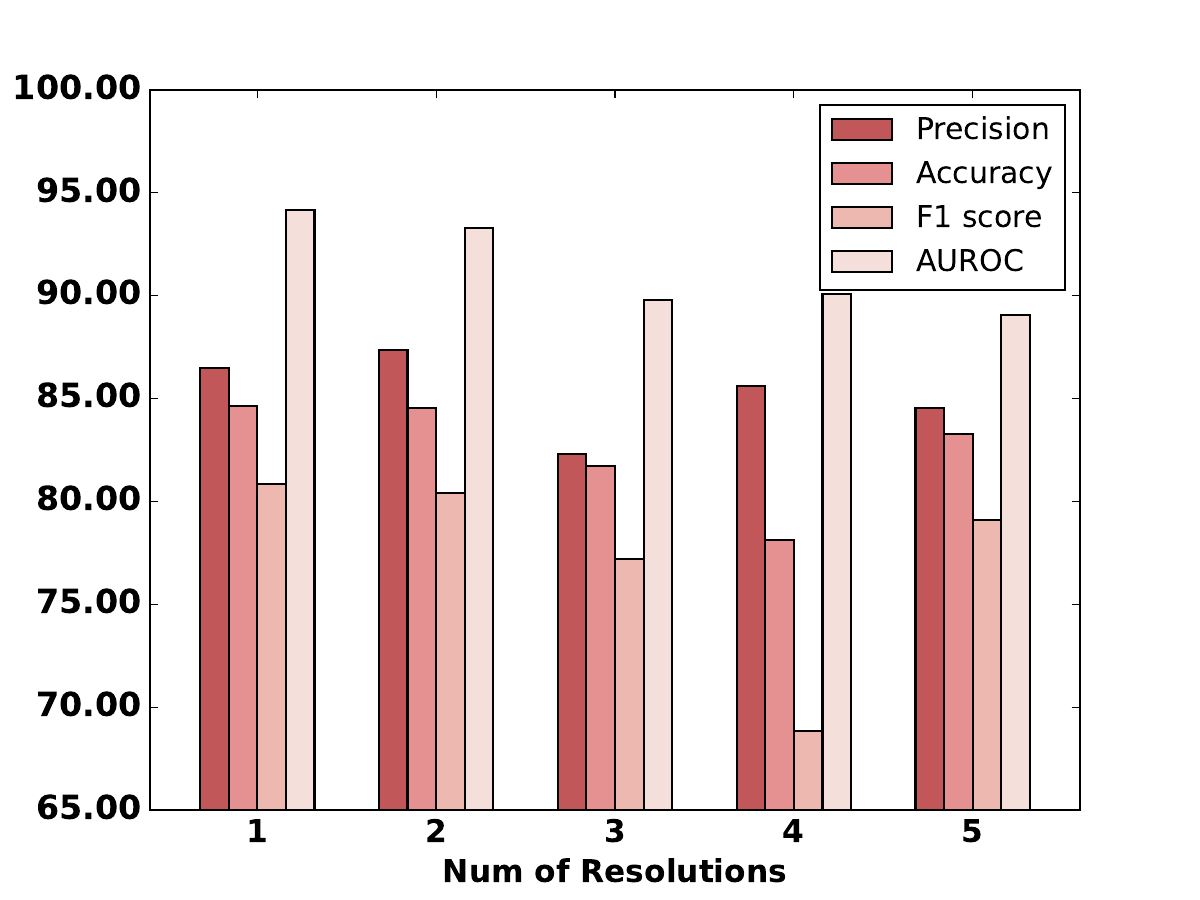}
    }
    \vspace{-4.5mm}
    \caption{Effect of different numbers of resolutions.}
    \label{fig:parameter_res_nums}
    \vspace{-5mm}
\end{figure}

\subsubsection{Efficiency Analysis}
We comprehensively evaluate the model efficiency from three aspects: classification performance (Accuracy), training speed, and memory footprint. Specifically, we choose two different sizes of datasets: the APAVA (23 subjects, 5,967 16-channel samples) and TDBRAIN (72 subjects, 6,240 33-channel samples) datasets. The closer a marker is to the upper-left corner of Figure \ref{fig:efficiency}, the higher the model's accuracy and the faster its training speed. Additionally, the smaller the marker’s area, the lower the memory usage during training. Therefore, we can conclude that although MedGNN's training time and memory footprint are at a moderate level among all baselines, its classification performance is the best. 

\subsubsection{Visualizations}
Figures \ref{fig:Adj_APAVA} and Figure \ref{fig:Adj_TDBRAIN} present visualizations of the learned adjacency matrices across different resolutions, providing insights into what the model captures at each resolution. Overall, the matrices tend to be sparse, indicating that MedGNN focuses on learning meaningful correlations from the data rather than relying on superficial variable aggregation. We also find that the weight distributions of adjacency matrices vary across different resolutions within the same dataset, indicating that the relationships between variables learned through graph learning differ at each resolution. This is significant for practical applications, as the relationships between variables are often not fixed but instead vary depending on the scale or context.

\subsubsection{Study of Numbers of Resolutions}
Figure \ref{fig:parameter_res_nums} shows the effect of the number of resolutions on classification performance. It suggests that appropriately increasing the number of resolutions has the potential to improve performance across various metrics. However, introducing overly coarse resolutions, where the receptive field becomes too large, may harm performance because excessive aggregation can lead to the loss of important fine-grained information and reduce the model's ability to capture subtle but critical patterns.

\begin{figure}[!t]
    \vspace{-4mm}
    \centering
    \subfigure[PTB-Subject]
    {
        \centering
        \includegraphics[width=0.47\linewidth]{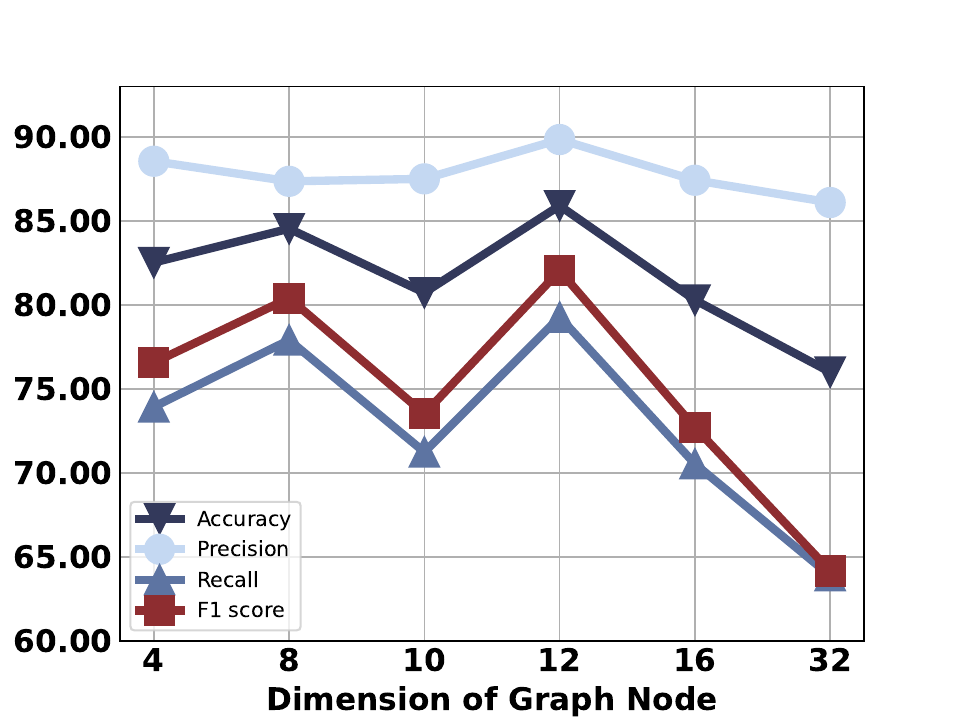}
    }
    \subfigure[APAVA-Subject]
    {
        \centering
        \includegraphics[width=0.47\linewidth]{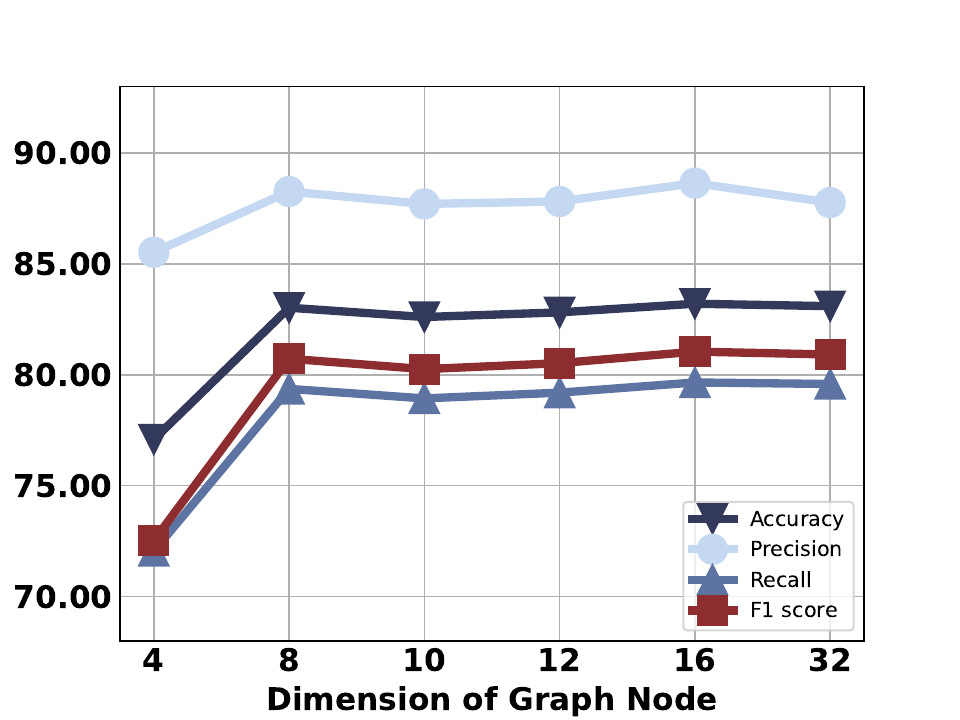}
    }
    \vspace{-4.5mm}
    \caption{Effect of different graph node dimension.}
    \label{fig:parameter_node_dim}
    \vspace{-5.5mm}
\end{figure}

\subsubsection{Study of the Size of Graph Node Dimensions.} Figure \ref{fig:parameter_node_dim} displays the effect of different graph node dimensions in multi-resolution graph learning. It indicates that larger graph node dimensions do not necessarily lead to better performance. This is understandable given that we employ a multi-resolution graph learning strategy, meaning the learning burden at each resolution is relatively light, making smaller node dimensions sufficient. In contrast, larger graph node dimensions may introduce data sparsity issues and increase computational overhead, hindering the model's learning ability. 

\vspace{-1.5mm}
\section{Conclusion}
In this paper, we have proposed a Multi-resolution Spatiotemporal Graph Learning framework, \textit{{MedGNN}}, 
for medical time series classification. We have constructed multi-resolution adaptive graph structures to learn dynamic multi-scale embeddings. Based on the graph structure, we have proposed two types of networks, i.e., Difference Attention Networks and Frequency Convolution Networks, to address the baseline wander problem in medical time series and learn the multi-perspective information for temporal modeling. We have also adapted Multi-resolution Graph Transformer for the dynamic spatial learning and information fusion. Extensive experiments have shown the superiority of our methods in different settings. We hope this could facilitate more future works on medical time series and healthcare applications.

\section{Acknowledgments}
This work was partially supported by National Natural Science Foundation of China (No. 62402531) and National Natural Science Foundation of China (No. 623B2043).

\bibliographystyle{ACM-Reference-Format}
\bibliography{sample-base}

\appendix
\section{Experimental Details}

\subsection{Datasets}\label{appendix_dataset}
The ADFD~\cite{miltiadous2023dataset} dataset includes 88 subjects and contains 69,762 multivariate EEG samples, each recorded from 19 channels. Every sample represents a one-second time sequence with 256 time points, captured at a sampling rate of 256Hz. Each sample is labeled with one of three classes, indicating whether the subject is Healthy, has Dementia, or has Alzheimer's disease.
The APAVA~\cite{escudero2006analysis} dataset includes 23 subjects and contains 5,967 multivariate EEG samples, each recorded across 16 channels. Each sample represents a one-second time sequence with 256 time points, recorded at a sampling rate of 256Hz. Every sample is accompanied by a binary label indicating whether the subject has been diagnosed with Alzheimer’s disease. 
The TDBRAIN~\cite{van2022two} dataset consists of 72 subjects and provides 6,240 EEG samples, each recorded across 33 channels. These samples represent one-second time sequences, containing 256 data points, recorded at a frequency of 256Hz. Each sample is tagged with a binary label that indicates whether the subject has Parkinson's disease. 
The PTB~\cite{physiobank2000physionet} dataset consists of 198 subjects and includes 64,356 multivariate ECG samples, each recorded across 15 channels. Each sample corresponds to a heartbeat represented by 300 time points, recorded at a sampling rate of 250Hz. A binary label accompanies each sample, indicating whether the subject has experienced a Myocardial Infarction.
The PTB-XL~\cite{wagner2020ptb} dataset comprises 17,596 subjects and includes 191,400 multivariate ECG samples, each recorded across 12 channels. Each sample represents a one-second time sequence with 250 time points, captured at a sampling rate of 250Hz. Each sample is labeled with one of five classes, representing different heart conditions.

\subsection{Baselines}\label{appendix_baseline}
We choose ten well-acknowledged and state-of-the-art models for comparison to evaluate the effectiveness of our proposed MedGNN for medical time series classification, including GNN-based models and Transformer-based models. We introduce these models as follows:

Medformer~\cite{wang2024medformer} introduces three innovative mechanisms to utilize the distinctive properties of medical time series. These include cross-channel patching for learning multi-timestamp and multi-channel features, multi-granularity embedding for capturing features at various scales, and a two-stage multi-granularity self-attention mechanism to capture features both within and across granularities. The official implementation is available at \url{https://github.com/DL4mHealth/Medformer}.

iTransformer~\cite{liu2023itransformer} reverses the structure of the Transformer by encoding individual series into variate tokens, allowing the attention mechanism to capture multivariate correlations, while applying feed-forward networks to each token for nonlinear representation learning. The official implementation is available at this repository: \url{https://github.com/thuml/iTransformer}.

PatchTST~\cite{nie2022time} segments time series into subseries-level patches, which serve as input tokens to the Transformer. These patch tokens replace traditional attention tokens, and a channel-independent structure further boosts efficiency. The official implementation is available at this repository: \url{https://github.com/yuqinie98/PatchTST}.

FEDformer~\cite{zhou2022fedformer} utilizes sparse attention with low-rank approximation in the frequency domain, achieving linear computational complexity and memory efficiency. It also introduces a mixture of expert decomposition to manage distribution shifts in time series~\cite{fan2023Dish}. The official implementation is available at this repository: \url{https://github.com/MAZiqing/FEDformer}.

Crossformer~\cite{zhang2022crossformer} embeds input data into a 2D vector array using Dimension-Segment-Wise embedding to retain time and dimension information, and employs a Two-Stage Attention layer to capture cross-time and cross-dimension dependencies efficiently. The official implementation is available at this repository: \url{https://github.com/Thinklab-SJTU/Crossformer}.

Autoformer~\cite{wu2021autoformer} uses an auto-correlation mechanism instead of self-attention and incorporates a decomposition block to separate trend and seasonal components, enhancing learning. The official implementation is available at this repository: \url{https://github.com/thuml/Autoformer}.

FourierGNN~\cite{yi2024fouriergnn} introduces a hypervariate graph, treating each series value as a node and representing sliding windows as space-time fully-connected graphs. It stacks Fourier Graph Operators (FGO) for matrix multiplications in Fourier space, providing high expressiveness with reduced complexity for efficient modeling. The official implementation is available at this repository: \url{https://github.com/aikunyi/FourierGNN}.

CrossGNN~\cite{huang2023crossgnn} refines both cross-scale and cross-variable interaction for multivariate time series with a linear complexity. Cross-scale GNN captures the scales with clearer trend and weaker noise, while cross-variable GNN maximally exploits the homogeneity and heterogeneity between different variables. The official implementation is available at this repository: \url{https://github.com/hqh0728/CrossGNN}.

TodyNet~\cite{liu2024todynet} captures hidden spatiotemporal dependencies without relying on a predefined graph structure. It introduces a temporal graph pooling layer to generate a global graph-level representation, leveraging learnable temporal parameters for graph learning. The official implementation is available at this repository: \url{https://github.com/liuxz1011/TodyNet}.

SimTSC~\cite{zha2022towards} frames time series classification (TSC) as a node classification problem on graphs. It introduces a graph construction strategy and a batch training algorithm with negative sampling to enhance training efficiency. The official implementation is available at this repository: \url{https://github.com/daochenzha/SimTSC}.

\subsection{Evaluation metrics}\label{appendix_evaluation_metrics}
To comprehensively and fairly evaluate the performance of each model in the classification task, we select five evaluation metrics: Accuracy, Precision, Recall, F1 score, and AUROC. The definitions and specific calculation formulas for each metric are presented below:

Accuracy measures the proportion of correct predictions out of the total number of predictions. It's calculated as:
\begin{equation}
    \text { Accuracy }=\frac{\text { Number of correct predictions }}{\text { Total number of predictions }}.
\end{equation}
This metric is useful when the classes are balanced but may be misleading in cases of class imbalance.

Precision focuses on the quality of positive predictions and measures the proportion of correctly predicted positive instances out of all instances predicted as positive. It’s especially useful when false positives need to be minimized. The formula is: 
\begin{equation}
    \text { Precision }=\frac{\text { True Positives }}{\text { True Positives }+ \text { False Positives }}.
\end{equation}

Recall measures the proportion of actual positive instances that were correctly identified. It’s important when false negatives are costly. The formula is:
\begin{equation}
    \text { Recall }=\frac{\text { True Positives }}{\text { True Positives }+ \text { False Negatives }}.
\end{equation}
It shows how well the model captures all relevant instances.

The F1 score is the harmonic mean of precision and recall, balancing the two when one is more important than the other. It’s particularly useful when dealing with imbalanced datasets, as it accounts for both false positives and false negatives. The formula is:
\begin{equation}
    \text { F1 Score }=2 \times \frac{\text { Precision } \times \text { Recall }}{\text { Precision }+ \text { Recall }}.
\end{equation}
It gives a single metric that reflects both precision and recall performance.

AUROC measures the model’s ability to distinguish between classes, regardless of the decision threshold. The ROC curve plots the true positive rate (recall) against the false positive rate (FPR), and AUROC is the area under this curve. A value of 1 indicates perfect classification, while 0.5 represents random guessing. It is a useful metric when the dataset is imbalanced and provides insight into how well the model separates the classes.

\subsection{Implementation Details}\label{appendix_implementation}
We follow the same data processing and train-validation-test set split protocol employed in Medformer~\cite{medformer_2024}. All the experiments are implemented in PyTorch 2.2.2~\cite{paszke2019pytorch} and conducted on a server equipped with four GeForce RTX 4090 GPUs, each with 24GB of memory. We utilize
ADAM~\cite{kingma2014adam} optimizer with an uniform initial learning rate $lr = 10^{-4}$ and cross-entropy loss for the model optimization. The batch size is selected from $\left\{32, 64, 128, 256\right\}$ and the number of training epochs is fixed to 10. We choose a subset from $\left\{2, 4, 6, 8, 10, 12, 14, 16\right\}$ as the multiple resolutions. The number of EncoderLayer $L$ is selected from $ \left\{4, 6\right\}$ and the dimension of embedding D is set from $\left\{256, 512\right\}$. And the graph node dimension is picked from $\left\{ 6, 8, 10\right\}$. We also report the standard deviation of MedGNN’s performance over five runs with different seeds, as shown in Table \ref{tab:sub-dep} and \ref{tab:sub-indep}, demonstrating the stability of its performance.

\end{document}